\documentclass[journal]{IEEEtran}

\usepackage{fancyhdr}
\usepackage{listings}

\ifCLASSINFOpdf
\else
\fi

\usepackage{amsmath}
\usepackage[url=false, sorting=none]{biblatex}
\usepackage{float}
\usepackage{graphicx}

\usepackage{algorithmic}
\usepackage{url}
\usepackage{hyperref}
\usepackage{subcaption}
\usepackage{graphicx}
\usepackage{amsfonts}
\usepackage{algorithm}
\usepackage{todonotes}
\usepackage{tcolorbox}
\usepackage{xcolor}
\usepackage{textcomp}
\usepackage{titlesec}
\usepackage{listings}
\usepackage{tabularx}
\usepackage{microtype}

\newcommand{\modify}[1]{{#1}} 

\definecolor{lightblue}{RGB}{173,216,230}

\definecolor{lightgray}{gray}{0.9}
\definecolor{darkgray}{gray}{0.3}
\definecolor{yaleblue}{rgb}{0.06, 0.3, 0.57}
\definecolor{dartmouthgreen}{rgb}{0.05, 0.5, 0.06}
\definecolor{steelblue}{rgb}{0.27, 0.51, 0.71}

\newtcolorbox{mybox}{
  colback=lightblue,
  colframe=lightblue,
  coltext=red,
  boxrule=0.5pt,
  arc=4pt,
  boxsep=2pt,
  left=2pt,
  right=2pt,
  before=\vspace{\baselineskip},
  after=\vspace{\baselineskip}
}

\let\originaltexttt\texttt
\renewcommand{\texttt}[1]{\textls[-30]{\originaltexttt{#1}}}

\tcbuselibrary{breakable} 

\newtcolorbox{prompt}[1][]{
  breakable,
  colback=white, 
  colframe=black, 
  colbacktitle=lightgray, 
  coltitle=black, 
  fonttitle=\bfseries, 
  title=#1, 
  fontupper=\usefont{T1}{ptm}{m}{n}, 
  rounded corners, 
  parbox=false,
  boxsep=5pt,
  parskip=1.5\baselineskip
}

\newtcolorbox{reply}[1][]{
  breakable,
  colback=lightgray, 
  colframe=black, 
  colbacktitle=darkgray, 
  coltitle=white, 
  fonttitle=\bfseries, 
  title=#1, 
  fontupper=\usefont{T1}{Roboto}{m}{n}, 
  rounded corners, 
}

\renewcommand{\markboth}[2]{}
\renewcommand{\markright}[1]{}

\begin{document}
\thispagestyle{fancy}

\title{Action Contextualization: Adaptive Task Planning and Action Tuning using Large Language Models}

\author{Sthithpragya Gupta$^{1,\dagger}$,
        Kunpeng Yao$^{1,\dagger,\ast}$,~\IEEEmembership{Member,~IEEE},
        Loïc Niederhauser$^{1}$,
        and Aude Billard$^{1}$,~\IEEEmembership{Fellow,~IEEE}
\thanks{{1}~Learning Algorithms and Systems Laboratory (LASA), School of Engineering, École Polytechnique Fédérale de Lausanne (EPFL), Switzerland}
\thanks{{$\dagger$}~Contributed equally.}
\thanks{{$\ast$}~Corresponding author.}
\thanks{\{sthithpragya.gupta, kunpeng.yao, loic.niederhauser, aude.billard\}@epfl.ch}
\thanks{This work was supported by the ERC Advanced Grant (ID: 741945, SAHR) and Horizon Europe (Grant No. 101070596, euROBIN).}
\thanks{The associated video is available at \url{https://youtu.be/6I9oQeWJUJ8};
experimental logs: \url{https://drive.google.com/drive/folders/1UFL8UkNH2m_RaaNimtdChK5SufWn2pk5};
Git repo:
\url{https://github.com/epfl-lasa/Action_contextualisation.git}.
}}

\maketitle
\thispagestyle{fancy}

\begin{abstract}
Large Language Models (LLMs) present a promising frontier in robotic task planning by leveraging extensive human knowledge. Nevertheless, the current literature often overlooks the critical aspects of robots' adaptability and error correction. This work aims to overcome this limitation by enabling robots to modify their motions and select the most suitable task plans based on the context.
We introduce a novel framework to achieve \emph{action contextualization}, aimed at tailoring robot actions to the context of specific tasks, thereby enhancing adaptability through applying LLM-derived contextual insights.
Our framework integrates motion metrics that evaluate robot performances for each motion to resolve redundancy in planning.
Moreover, it supports online feedback between the robot and the LLM, enabling immediate modifications to the task plans and corrections of errors.
An overall success rate of $81.25\%$ has been achieved through extensive experimental validation.
Finally, when integrated with dynamical system (DS)-based robot controllers, the robotic arm-hand system demonstrates its proficiency in autonomously executing LLM-generated motion plans for sequential table-clearing tasks, rectifying errors without human intervention, and showcasing robustness against external disturbances.
Our proposed framework also features the potential to be integrated with modular control approaches, significantly enhancing robots' adaptability and autonomy in performing sequential tasks in the real world.
\end{abstract}


\IEEEpeerreviewmaketitle


\section{Introduction}
Task and Motion Planning (TAMP) in robotics aims to devise motion sequences that guide a robot toward a predetermined goal, integrating task planning with motion planning. Effective TAMP requires environmental information to ensure coherent, achievable, and constraint-compliant plans.
Traditional task planning relies on comprehensive environmental descriptions and substantial computational resources, often using languages like PDDL (Planning Domain Definition Language) or STL (Signal Temporal Logic). Such model-based approaches often require manual input, such as domain files and problem files for PDDL or signal descriptions and temporal logic formulas for STL, restricting the autonomy of the robot and the feasibility of solutions when facing unknown environments.

Large Language Models (LLMs) have recently shown promise in various robotic tasks, offering a new avenue for TAMP solutions.
The recent advancement of LLMs offers a promising solution for tackling TAMP problems in robotics \cite{misra_tell_2016,ahn2022can,singh2023progprompt,jin2024robotgpt}.
LLMs can leverage extensive human knowledge to generate robot task plans \cite{ding2023task,zhou2023isr} or directly program robots \cite{liang2023code}, alleviating the need for explicit manual modeling of detailed task information.
However, LLMs can produce unreliable outputs, particularly in ensuring the feasibility on real robots, due to their black-box nature. This gap between conceptual plans and physical execution can hinder task feedback integration, leading to failures. Hence, LLM-based approaches for robots are often limited to simple pick-and-place tasks in simulations, struggling with long-horizon manipulations, in particular, the ones involving physical interactions.
To address these challenges, LLMs must generate logical task plans that are both feasible for robots and adaptive to task contexts, satisfying explicit and implicit constraints. We refer to this capability as \emph{action contextualization}.
For this purpose, we present a framework that allows for tuning parameters of robotic actions according to the inferred task context. The framework also evaluates motion viability through simulations and defined metrics, resolving redundancies in planning.

Our main contributions are threefold:
\begin{itemize}
    \item First, we elaborate on a nuanced approach for tuning action parameters, such as velocity and orientation, enhancing the adaptability of motions to the specific demands of each step of the task;
    \item Second, we leverage feedback for \emph{motion tuning} and strategic \emph{task replanning}, and employ a \emph{motion evaluation} scoring system to select the optimal plan;
    \item Third, we integrate LLM-based planning with an online collision detection approach and DS-based robot controllers, significantly enhancing robots' robustness and adaptability in the real world.
\end{itemize}

\section{Background}
\subsection{Task and Motion Planning}
Traditional methods for TAMP often rely on hierarchical planning, integrating high-level task planning with low-level motion planning \cite{garrett2020pddlstream}.
\cite{pflueger2015multi} introduced multi-step planning as a solution for TAMP in robotics, demonstrating its effectiveness with a dual-armed robot handling a complex object.
\cite{gorner2019moveit} presented the Task Constructor framework in \emph{MoveIt!} \cite{chitta2012moveit}, enhancing the capabilities of robotic manipulation by allowing the definition and planning of motions consisting of multiple sub-tasks.
While symbolic knowledge is crucial in task planning, motion planning often involves continuous models of robot trajectories or behaviors, presenting a significant challenge in integrating symbolic and continuous paradigms.
Sampling-based methods and trajectory optimization have been used to address this issue by turning task planning into a sampling process within a probabilistic search space. However, this approach may not always be effective in finding solutions during complex task-level searches \cite{mansouri2021combining}.
\cite{jiang2019task} proposed TMP-RL, a framework integrating TAMP and reinforcement learning (RL) to enhance robustness in planning across dynamic and uncertain domains, thereby improving adaptability to previously unseen domain dynamics.
\cite{kast2020hierarchical} developed a system with a hierarchical planner that solves TAMP problems while optimizing for parallelization. This system uses action models based on a constraint formulation, allowing asynchronous parallel execution of plans on dual-arm robots.
However, these traditional TAMP approaches often exhibit adaptability, generalization, and computational efficiency limitations.
This is mainly due to their reliance on predefined models and scenarios, which limits their ability to adapt to new or complex environments, and the high computational demands of planning in detailed, dynamic spaces.
Moreover, many of these approaches do not guarantee completeness, potentially overlooking infeasible problems.
We address this issue by defining a set of actions and their desired features based on robots' capabilities without requiring explicit environmental modeling. This allows the framework to be generalized to similar robots, novel tasks, and unknown environments.

\subsection{LLMs for TAMP in Robotic Tasks}
With their ability to generalize and reason about various domains, LLMs have demonstrated potential as task-agnostic planners \cite{sharma2021skill,zeng2022socratic}, showcasing TAMP capabilities in various robotic tasks, such as pick-and-place \cite{singh2023progprompt} and object rearrangement \cite{ding2020task}.
Efforts to improve the effectiveness of LLMs include incorporating human feedback to align better planning with user intent \cite{ouyang2022training}.
In \cite{wu_tidybot_2023}, the applicability of LLMs on robots for tidying up a room, using commonsense knowledge and a limited action set, was investigated.
Incorporating domain knowledge, LLMs have also been used as generalized planners in PDDL domains \cite{garrett2020pddlstream, silver_generalized_2023}.
Moreover, to enhance LLM-based robotic task planning performance, \cite{zhou2023isr} proposed a validation framework through iterative self-refinement to improve task success rates.
\cite{zha2023distilling} proposed DROC, handling various language feedbacks and retrieving past experiences, but relies heavily on human corrections, which can be subjective, time-consuming, and inconsistent. Our framework adapts by extracting task-specific contextual information and tuning/replanning robotic actions in simulation.
More recently, \cite{jin2024robotgpt} proposed RobotGPT to improve task success by leveraging ChatGPT's problem-solving insights within a stable framework, addressing limitations, such as unpredictability and safety concerns.
Despite the promising advancements, there remains a considerable gap in fully integrating these advanced planning capabilities with real robotic applications. Challenges persist in ensuring these approaches' safety, interpretability, and real-world applicability.
In this paper, we demonstrate that this gap can be bridged by employing dynamical system-based adaptive robot controllers for low-level motion control, ensuring adaptability and robustness while executing the high-level task plan.

\subsection{Action Grounding}
A key aspect of language-aided robotic task execution is grounding language commands into executable robot actions \cite{glenberg_grounding_nodate}. 
Operator grounding, i.e., connecting abstract, language-based task instructions to their practical execution by robots, serving as a subset of the broader symbolic grounding problem, establishes the link between the abstract language-based task instructions and their practical execution on the real robots.
Specific models such as \cite{tellex_understanding_2011, howard_natural_2014, liu_jointly_2016} can determine connections between elements of the language instructions and the semantic entities in the environment.
However, they often do not include the underlying relational knowledge, which refers to understanding how different concepts, objects, or entities relate to each other within a context in their reasoning process, potentially limiting the depth of comprehension and interaction with the environment.
The specific context of a task significantly influences the physical actions undertaken by the actual robots~\cite{misra_tell_2016}.
Crucially, it's essential to effectively link the LLM's abstract instructions to the robot's concrete actions to ensure successful task completion.
Contemporary works have started leveraging LLMs to achieve action grounding by guiding the agent's interaction with the environment \cite{carta2023grounding}.
For example, low-level pre-trained skills combined with LLMs can facilitate knowledge acquisition and task execution \cite{ahn2022can}.
\cite{liu_lang2ltl_2023} proposed an approach to translate abstract instruction into formal logic that the robots can interpret. In contrast, \cite{lin2023text2motion} assessed the geometric feasibility during the robot's action selection. 
\cite{ding2023task, huang2022language} utilized prompting to extract the context established by the task and identify the admissible actions for the robot to execute.
Action grounding with LLMs still faces challenges such as satisfying real-world physical constraints, ensuring the feasibility for robots, and bridging the gap between planned and actual outcomes. These issues highlight the difficulty in translating abstract LLM instructions into tangible robotic actions, indicating the need for improved methodologies.
We address these challenges by implementing a prior validation, including motion tuning and task replanning to verify constraint satisfaction and action feasibility. We also employ online collision detection and DS-based robot controllers to ensure adaptive and robust execution.


\section{Method}
\label{sec:method}

\begin{figure*}[htp!]
    \centering
    \includegraphics[width=\linewidth,trim={0 0 0 0},clip]{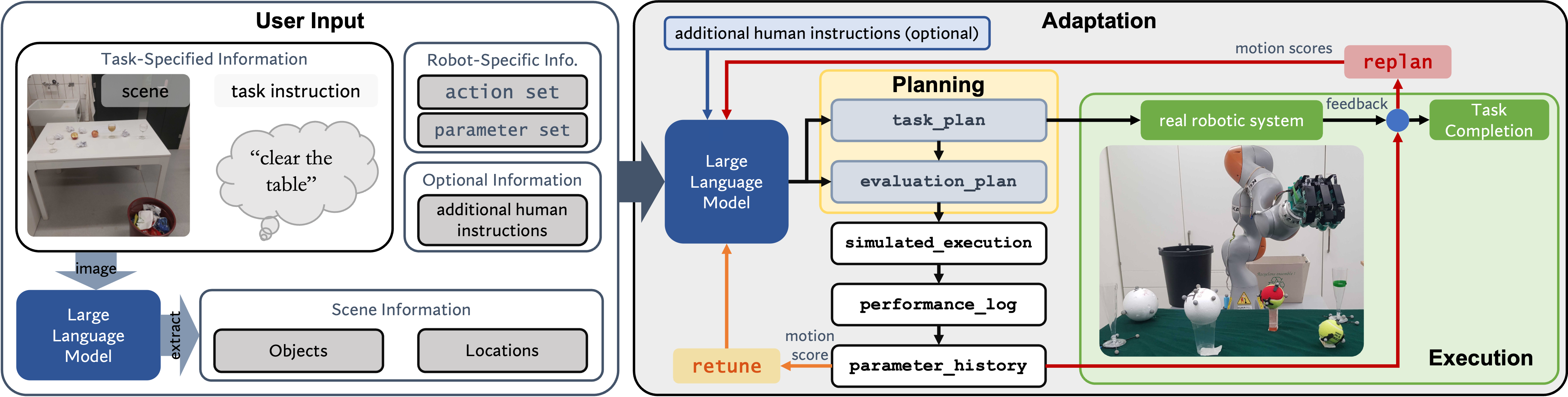}
    \caption{The proposed framework for performing context-aware task planning and action parameter tuning for robotic sequential tasks. The red arrow lines indicate the \texttt{retune} or \texttt{replan} of the parameters, while the grey lines indicate the process for the real robotic system to execute the task once the action plan is tuned}\label{fig:single_format_framework}
\end{figure*}

We propose a framework to achieve action contextualization (see Fig.~\ref{fig:single_format_framework}). The framework initializes by feeding the instruction and a scene image of the task to the LLM to generate a description, establishing context. The LLM extracts \texttt{objects} and \texttt{locations} lists from the scene and description, \modify{that provide object labels used in planning}. \modify{We symbolically denote the Python variable or method declared in the framework as \texttt{variable}, in the following body.}
We also provide the task to be performed by the robot. 
\modify{Using the description, task, objects, and locations, the framework next undertakes a three-step procedure to real-world robot execution:}
\begin{enumerate}
    \item \textbf{Planning}: the framework prompts the LLM to generate a task plan using the robot's available motions with initial parameters;
    \item \textbf{Adaptation}: the plan is simulated, and the performance feedback is used to tune action parameters, with replanning if needed;
    \item \textbf{Execution}: the optimized plan is executed by the real robotic system.
\end{enumerate}

\subsection{Planning}\label{subsec:planning}
\textbf{Task plan} The first step towards real-world task execution starts with preparing a plan consisting of a sequence of actions for the robot. 
This plan, formulated as a Python list of the actions' symbolic expressions (listed in Table \ref{tab:actions}), is hereby referred to as the \texttt{task\_plan}.

The LLM is prompted to generate this list, only utilizing the actions in the
repertoire (Table~\ref{tab:actions}) to ensure the plan does not exceed the
robot’s movement ability.

Each action is parameterized by a combination of variables, including (1) qualitative parameters, which are symbolic expressions, such as the desired \texttt{object}, \texttt{location}, and \texttt{grasp}, and (2) quantitative parameters, which are numeric values, such as \texttt{speed}, \texttt{orientation}, and \texttt{obstacle\_clearance}.
The qualitative parameters regulate the action sequences in the \texttt{task\_plan}, whereas the quantitative parameters regulate the robot behavior (see Table \ref{tab:parameters}).

The LLM initializes the \texttt{task\_plan} with an initial estimation of actions and their associated parameters and tunes them over subsequent iterations. This is further explained in Sec.~\ref{subsec:tuning}.

\begin{figure}
    \centering
    \includegraphics[trim=0cm 0 0 0,clip,width=\linewidth]{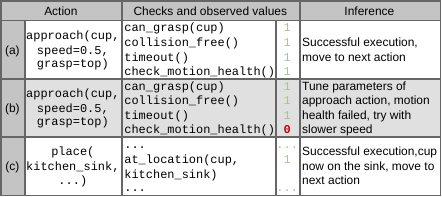}
    \caption{\modify{Instances of actions from task plan and the corresponding checks from evaluation plan. Observations from the checks provide useful inference to LLM: (a) points to safe, collision-free and timely execution of approach, resulting in the robot being close enough to grasp the cup; (b) the action failed due to a false \texttt{check\_motion\_health()}, which can be overcome by tuning the motion by using a lower speed value; (c) the action was successful and the cup is now on the kitchen sink}}
    \label{fig:eval_plan_instances}
    \vspace{-\baselineskip}
\end{figure}

\textbf{Evaluation plan} To evaluate the outcome of each action in the \texttt{task\_plan}, the framework next prompts the LLM to generate an \texttt{evaluation\_plan} which consists of various \textit{checks} and expected outputs to determine whether the execution of an action in the \texttt{task\_plan} is successful or not.
These checks, listed in Table \ref{tab:checks}, serve as predicate classifiers - functions that map continuous variables such as the state of the robot and (or) its environment to truth values. 
The desired truth values corresponding to the predicate classifiers contained in the checks describe the intended effect of the action. 
Comparing the observed truth values generated by these checks during action execution with their desired counterparts determines the success of the action (see Fig.~\ref{fig:eval_plan_instances} for examples).

For instance, in Fig.~\ref{fig:eval_plan_instances}(a), when executing the \texttt{approach(cup)} action, we expect to observe \texttt{can\_grasp(cup)} = 1, i.e., the robot has reached close-enough to the grasp the \texttt{cup}; \texttt{collision\_free()} = 1, i.e., the robot's motion was collision free; \texttt{timeout()} = 1, i.e. the action was executed in a timely fashion; \texttt{check\_motion\_health()} = 1, i.e., the robot's motion was safe for its hardware.
Similarly, if the \texttt{can\_grasp(object\_to\_grasp)} check yielded false post \texttt{approach(cup)}'s execution, it would signify that the action failed.   

The LLM is prompted to generate the \texttt{evaluation\_plan}, a Python list of relevant checks for each action, using those listed in Table \ref{tab:checks}.
Of these checks, it is mandatory to use \texttt{collision\_free()}, \texttt{timeout()}, and \texttt{check\_motion\_health()} for each action. The use of the remaining checks and their judgment of applicability are left to the discretion of LLM.
The \texttt{evaluation\_plan} serves the following purposes:
\begin{enumerate}
    \item Providing feedback to the LLM on the state of the environment between actions. For instance in Fig.~\ref{fig:eval_plan_instances}(c), to place the \texttt{cup} on the \texttt{kitchen\_sink}, the \texttt{place(kitchen\_sink,...)} action is used. By following this up with the \texttt{at\_location(cup, kitchen\_sink)} check, we can not only check the success of \texttt{place}, but also convey to the LLM that the \texttt{cup} is now on the \texttt{kitchen\_sink}.
    \item The check(s) that failed enable the LLM to formulate hypotheses on why the action failed and undertake appropriate adaptation (more in Sec.~\ref{subsec:tuning}). For instance, in Fig.~\ref{fig:eval_plan_instances}(b), if \texttt{approach(cup, speed=0.5, grasp=top)} failed due to \texttt{check\_motion\_health()} being false, the LLM may hypothesize that the robot's motion was too fast, which could have compromised the hardware integrity of the robot. Hence, trying with a lower speed value could be a plausible approach.
    \item Precisely identifying at which step the \texttt{task\_plan} encountered failure and the unsuccessful action. This introduces a chronological perspective when the LLM tries to understand the sequence of events leading to the action's failure. For instance, consider a \texttt{task\_plan} that directs the robot to perform a pick-and-throw motion for a collection of 3 objects lying on a table - \texttt{used\_tissue}, \texttt{rotten\_apple}, and \texttt{rubber\_ball} - in that order to land these items in the \texttt{trashbin}. During execution, when dealing with the \texttt{rubber\_ball}, say the \texttt{drop(trashbin)} action successively encounters failure despite adaptation. Thus, when undertaking replanning (more in Sec.~\ref{subsec:tuning} and \ref{sec:execution}), the LLM recycles the first successful steps when dealing with \texttt{used\_tissue} and \texttt{rotten\_apple}, only when interacting with \texttt{rubber\_ball}, would the LLM alter the \texttt{task\_plan} to try \texttt{place(trashbin)} instead of dropping it from height.
\end{enumerate}

With every iteration, the automated feedback generated through the \texttt{evaluation\_plan} (which checks failed) is utilized by the framework to progressively tune the action parameters.

\begin{table*}
    \centering
    \caption{List of the actions that the robot can execute}
    \begin{tabular}{ll}
         \hline
         Action & Description \\
         \hline
         \texttt{drop(location, speed, obstacle\_clearance)} & Drops the grasped object over the \texttt{location}\\
         \texttt{place(location, orientation, speed, obstacle\_clearance)} & Positions the grasped object at the \texttt{location} and\\
         & release the grasp\\
         \texttt{pick(object\_to\_grasp, speed, obstacle\_clearance, grasp)} & Instructs the robot to pick up the \texttt{object\_to\_grasp}\\ 
         & if it is close enough\\
         \texttt{approach(object\_to\_grasp, speed, obstacle\_clearance, grasp)} & Moves the robot close to \texttt{object\_to\_grasp} so that\\
         & the object is in the robot's reach\\
         \hline
    \end{tabular}
    \label{tab:actions}
\end{table*}

\begin{table*}
    \centering
    \caption{List of action parameters and descriptions}
    \begin{tabularx}{\textwidth}{llX}
         \hline
         Parameter & Type & Description\\
         \hline
         \texttt{object\_to\_grasp} & Symbolic & Label of the object that the robot interacts with\\
         \texttt{location} & Symbolic & Label of the goal location for the object grasped by the robot\\
         \texttt{speed} & Numeric & A value in $[0,1]$ that regulates how fast the robot moves. A value closer to 1 instructs the robot to move at high speed, but this may result in a jerky motion\\
         \texttt{orientation} & Numeric & A value in $[0,1]$ that regulates how crucial it is for the robot to maintain the original orientation of the object that has been grasped. A value closer to 1 instructs the robot to maintain the orientation strictly but may result in difficulty in avoiding external perturbations\\
         \texttt{obstacle\_clearance} & Numeric & Specifies how close the robot can get (in meters) to an object (including the one it is trying to grasp) before starting to avoid it. Small values allow the robot to get closer to obstacles and usually give a better chance of reaching the object, being able to grasp them, or placing them in the correct place. However, small values also increase the chance of getting into a collision. The reasonable range is between 0 and 0.1 meters; typical values are 0.005 to 0.02 meters\\
         \texttt{grasp} & Symbolic & One of the two values from \{\texttt{top}, \texttt{side}\}. \texttt{top} instructs the robot to approach the object, to grasp it from the top; \texttt{side} instructs the robot to approach the object from the side\\
         \hline
    \end{tabularx}
    \label{tab:parameters}
\end{table*}

\begin{table*}
    \centering
    \caption{List of the available checks for evaluating actions' execution. For more details, see Appendix~\ref{app:checks}}
    \begin{tabularx}{\textwidth}{llX}
         \hline
         Check & \modify{Applicability}\quad\quad\quad & Description\\
         \hline
         \texttt{collision\_free()} & \modify{Mandatory} & Detects collisions during the preceding action's execution\\
         \texttt{timeout()} & \modify{Mandatory} & Ensures the preceding action was executed within a reasonable time\\
         \texttt{check\_motion\_health()} & \modify{Mandatory} & Evaluates the safety of the robot during the execution of the preceding motion\\
         \texttt{can\_grasp(object\_to\_grasp)} & \modify{LLM's discretion} & Verifies if robot is within grasping distance of \texttt{object\_to\_grasp}\\
         \texttt{holding()} & \modify{LLM's discretion} & Verifies if the robot has grasped an object\\
         \texttt{at\_location(object, location)}\quad\quad\quad & \modify{LLM's discretion} & Confirms if \texttt{object} is at \texttt{location}\\
         \texttt{can\_reach(goal)} & \modify{LLM's discretion} & Assesses feasibility of robot reaching \texttt{goal} object or location from current state\\
         \hline
    \end{tabularx}
    \label{tab:checks}
    \vspace{-\baselineskip}
\end{table*}

\subsection{Adaptation}
\label{subsec:tuning}

Once the \texttt{task\_plan} and \texttt{evaluation\_plan} are initialized, the framework starts to iteratively tune the action parameters during the simulated task execution. This is carried out using the LLM, in accordance with the task context.
Finally, the most suited \texttt{task\_plan} and associated \texttt{evaluation\_plan} are selected for the robot to perform the task (see Fig.~\ref{fig:single_format_framework}).
\subsubsection{Iterative motion adaptation}
\label{subsubsec:motion_adaptation}
The initialized \texttt{task\_plan} undergoes simulation-based validation, where the \texttt{evaluation\_plan} determines the expected/ideal behavior and provides automated feedback, which includes:
\begin{itemize}
    \item Failed step of the \texttt{task\_plan}, i.e., the unsuccessful action;
    \item Reason of failure: determined by the failed check-in \texttt{evaluation\_plan};
    \item A motion score associated with each action executed until failure (see Sec.~\ref{subsubsec:motion_score}).
\end{itemize}
The feedback is stored as \texttt{performance\_log}.

In addition, the parameters of both successful and unsuccessful actions (and the corresponding objects being manipulated) are also logged in \texttt{parameter\_history}.
The feedback informs regarding the failed action, whereas the log of successful and failed parameters for the action is used to inform the tuning process. Post tuning, the resulting \texttt{task\_plan} undergoes validation anew. The resultant of the execution is used to update the \texttt{performance\_log} and append to the \texttt{parameter\_history}. This process constitutes one cycle of the \texttt{retune} process.
The \texttt{retune} process continues until the framework converges to a tuned \texttt{task\_plan}, or a certain amount of iteration has been conducted.

If the tuned \texttt{task\_plan} remains unsuccessful, the \texttt{replan} procedure prompts the LLM
to assess feedback from the \texttt{evaluation\_plan} and revise
the action sequence in the task plan. 
During this procedure, adjustments may be made to various aspects of the \texttt{task\_plan}, including the object-interaction sequence (e.g., interacting first with \texttt{obj2} instead of \texttt{obj1}); action parameters; the type of actions (e.g., using \texttt{drop} instead of \texttt{throw}).

Once the \texttt{task\_plan} passes simulation-based validation, it will be applied to guide the real robotic system to perform tasks in the real world.

\subsubsection{Motion evaluation}\label{subsubsec:motion_score}
Multiple motions could be feasible to achieve the desired effect(s) at a step of the plan due to redundancy.
For example, the robotic arm can reach a target object via multiple feasible trajectories and grasp the object from different directions, even using different hand poses.

To resolve the redundancy in generated motions, we evaluate sets of feasible motions across multiple independently generated plans and use the one having the highest \emph{quality score}, $\mathcal{Q}$, which considers both internal and external metrics.

The internal quality metric $\mathcal{M}_{I}$ evaluates the quality of robot configuration during the motion.
The dexterity index \cite{klein1987dexterity}, defined as ${\mu_{max}}/{\mu_{min}}$ with $\mu_{max}$ and $\mu_{min}$ being the largest and smallest singular values of the manipulator Jacobian matrix $J$, respectively.
It indicates the ``uniformity'' of the manipulability ellipsoid \cite{yoshikawa1985manipulability}.
We use the inverse of the original dexterity index, $\delta = {\mu_{min}}/{\mu_{max}}$, so that its value is normalized in $(0,1]$.
The ``ideal'' configuration has a value $\delta=1$, indicating an \emph{isotropic} configuration \cite{salisbury1982articulated}, where the end-effector can freely move in all directions in the space.
The index value approaches $0$ as the robot moves towards a singularity, i.e., a configuration with impaired motion ability in some directions, hence being unable to execute subsequent motions.
We keep track of the metric values during the entire motion trajectory, and the smallest value recorded is assigned as the internal score of the motion: $\mathcal{M}_{I} = \min\{\delta\}$.

The external quality metric $\mathcal{M}_{E}$ evaluates the safety of the robotic system, measured by its minimal distance to the environment during the motion, $\mathcal{M}_{E} = \min\{d_{i}\}/\bar{d}$, $i=1,\dots, N_{L}$, where $d_{i}$ is the minimal distance of the $i$th link to its environment, measured by the neural distance function;
$\bar{d}$ is the maximum distance from any robot link to the workspace's border (the region of interest). $N_{L}$ is the total number of links in the robotic system.
The total motion quality score is calculated as $\mathcal{M}_{T} = \mathcal{M}_{I} + \mathcal{M}_{E}$.

Moreover, it is worth noting that this evaluation approach can be used to select the best task plan among multiple feasible ones for a specified task.
To this end, we evaluate each generated plan by calculating the expectation of the motion scores along this plan, $\mathcal{Q} = \mathbb{E}[\mathcal{M}_{T_{s}}]$, $s=1,\dots, N_{s}$, where $N_{s}$ is the total number of motions in the plan.
The plan with the highest score guides the real robot in performing the sequential task.

\subsection{Execution}
\label{sec:execution}
Finally, after Planning (Sec.~\ref{subsec:planning}) and Tuning (Sec.~\ref{subsec:tuning}), the most effective plan for accomplishing the specified task is identified.
To ensure real-world execution efficacy, the framework must be robust to failure during robot execution and incorporate additional feedback from a human expert.

The simulated success of the finalized \texttt{task\_plan} may not necessarily carry on to the real world. It is required to plan for failure alleviation.
When faced with failure, the \texttt{replan} procedure is promptly called.
Contrary to the \texttt{retune} procedure implemented during the tuning process in Sec.~\ref{subsec:tuning}, during real-world execution, the \texttt{replan} procedure leverages feedback from the \texttt{evaluation\_plan} to alter the \texttt{task\_plan} from the current step. This allows the agent to correct the failure induced during the previous unsuccessful action before proceeding with the alteration and (or) execution of the remainder of the steps. For instance, a failed \texttt{pick(apple)} would require the agent first to execute \texttt{approach(apple)} before retrying \texttt{pick(apple)}.
In addition to using the feedback from \texttt{evaluation\_plan}, both \texttt{replan} and \texttt{retune} are capable
of using additional external feedback during execution, such
as from a human expert or new sensory information.


\section{Experiments}
In this section, we conduct experiments to assess our proposed approaches.
We first describe the experimental task and system specification in Sec.~\ref{subsec:exp_setup}.
In Sec.~\ref{subsec:exp_sce_1}, we evaluate the framework's capability for tuning parameters for a sequence of actions. Sec.~\ref{subsec:exp_contextualization} highlights the framework's context-aware adjustment ability by demonstrating how the framework selects appropriate actions (and parameters) according to the object and subsequent manipulation goals by leveraging the prior human knowledge encapsulated in the LLM.
Sec.~\ref{subsec:exp_sce_2} extends this evaluation to tuning parameters across an entire task plan, demonstrating the LLM's effectiveness in task planning and action selection.
In Sec.~\ref{subsec:LLM_comparison}, we test different LLM models to perform the same task.
Next, in Sec.~\ref{subsec:random_experiments}, we test the robustness of the framework to instances of randomization in the environment.
Finally, in Sec.~\ref{subsec:exp_sce_3}, we proceed with the real-world robotic execution of the best task plan generated by the framework.

\subsection{Experimental Setup}\label{subsec:exp_setup}

\subsubsection{Task description}
\begin{figure}[]
    \centering
    \includegraphics[trim=0cm 0 0 0,clip,width=\linewidth]{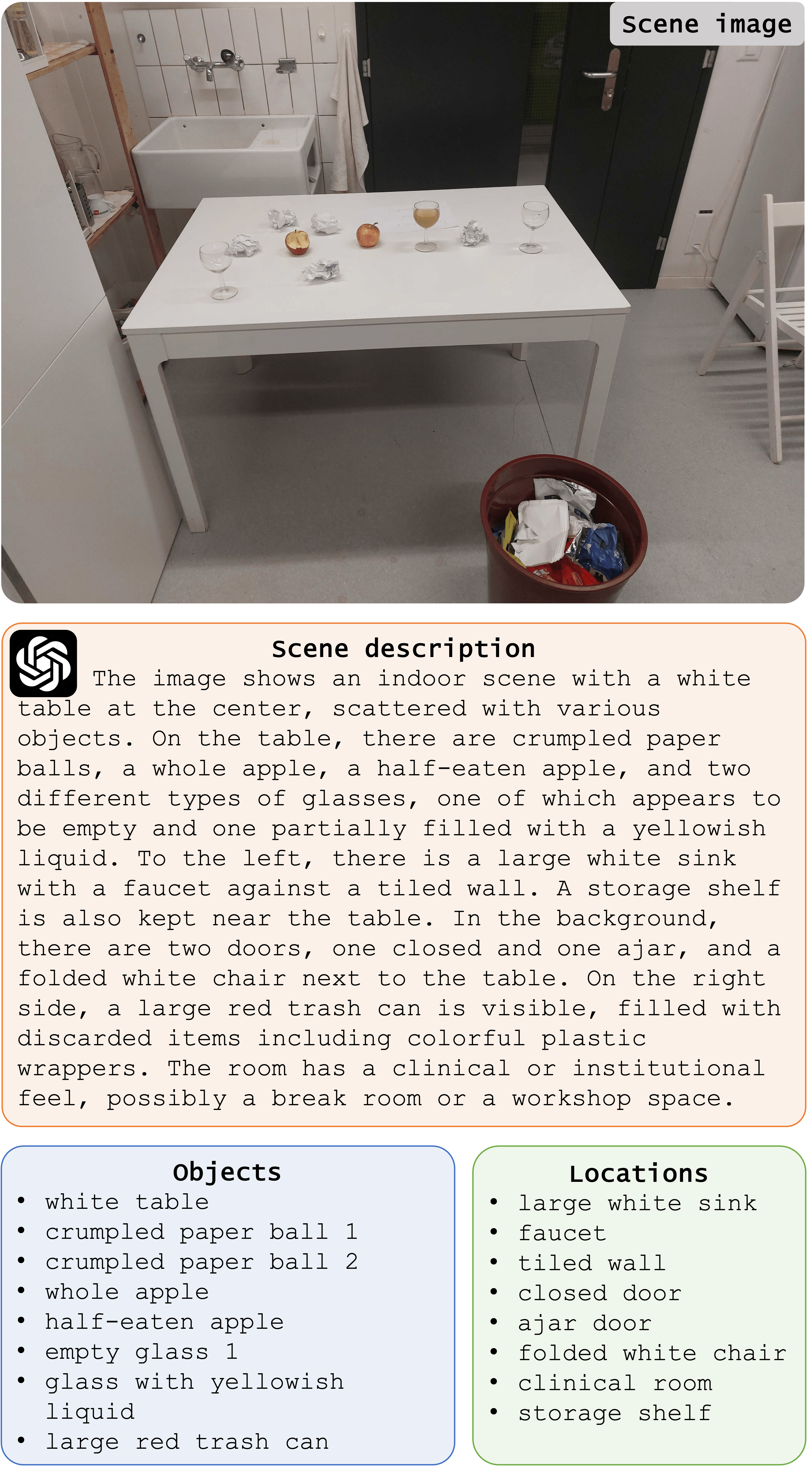}
    \caption{The input scene and description generated by LLM, accompanied by the extracted objects and locations}
    \label{fig:scene_description}
\end{figure}

We consider a table-cleaning task to evaluate our framework: a robotic arm-hand system is tasked to efficiently clean the surface of a messy dining table with various objects placed on it, including a whole apple, a half-eaten apple, a full champagne glass, an empty champagne glass, and five paper balls (see Fig.~\ref{fig:scene_description}). Next to the table is a trash can to throw residual food and a sink to wash dishes.
The corresponding environment in simulation is presented in Fig.~\ref{fig:retune_process}, where the only difference is we simulate two paper balls instead of all five.
In such a long-horizon sequential task, the robot must (1) generate a sequence of tackling the cluster of objects and (2) automatically choose the correct motion and set a goal for each object, respectively.
In addition to logically performing a sequence of motion, the robot needs to closely interact with the objects, such as grasping and polishing.
Moreover, such motions must be performed in a crowded environment, increasing the task difficulty.

\subsubsection{System specification}
We used a robotic setup with a 7-DoF KUKA iiwa arm equipped with a 16-DoF Allegro hand as the end-effector.
This setup is used in both simulations and real robotic experiments.
We used MuJoCo\footnote{\url{https://mujoco.org/}} as a physics engine for simulation.
We use \texttt{GPT-4-1106-preview} as our LLM engine in this paper. For comparison between different LLM models in Sec.~\ref{subsec:LLM_comparison}, we additionally use the \texttt{open-mixtral-8x22b} and \texttt{Meta-Llama-3-70B} models.
For simplification, we use the term ``LLM'' in this paper. However, notice that GPT-4 is essentially a commercial product of the OpenAI company.

We generate trajectory in joint space with a linear dynamical system (DS) to guide the arm towards the goal while avoiding spatial obstacles.
The robotic arm uses a computed torque control with joint space DS modulation \cite{huber2022fast} for collision avoidance. This modulation is constructed using the distances estimated by a joint neural distance function \cite{chen2023differentiable}. A joint position controller controls the hand to achieve open and close poses.
The robotic system communicates with the controller PC via the Robotic Operation System (ROS) at 200 Hz.
The controller PC has an Intel i9-10900K CPU (3.70GHz) and an NVIDIA GeForce 2080 Ti GPU.
Moreover, in the real robotic experiment, we put markers on experimental objects and use an OptiTrack\textsuperscript{\texttrademark} motion capture system to track the position and orientation of objects in real time.

\subsubsection{Context for the task}
The LLM first generates a description of the scene and extracts a list of possible objects and locations before the experiments (see Fig.~\ref{fig:scene_description}). The scene descriptions establish the context for the planning problem, and the extracted objects and variables serve as parameters for actions and checks.
The framework prompts the LLM to generate a scene description and extract possible objects and locations as Python lists before the experiments. These supply the object labels used by actions and checks. To increase reproducibility, the detailed scene description generated by the LLM and all prompts used in our experiments are available in the experimental logs and the appendices.

\subsection{Evaluation of Action Tuning}\label{subsec:exp_sce_1}
In this experiment, we evaluate the ability of our framework to tune the action parameters.
The objects were randomly placed on the table surface at the beginning of the experiment.
Next, six tasks - one per object - are assessed independently. 
A total of $8$ trials per task are performed. This allows us to assess action tuning in response to each object.
In each trial, we select the relevant object from the table and then prompt the LLM to generate a suitable motion plan to clear it, i.e., throwing it into the trash bin or the sink or transferring it to a different place.
The robotic system then executes the motion plan in the simulation (see Fig.~\ref{fig:single_format_framework}).
Once a motion plan is generated and executed, we inform the LLM of the scores (both internal and external scores) and prompt it to tune the parameters of each action in the plan to improve the overall motion score.
The LLM iterates the retune process four times.
Then, we prompt it to re-generate the motion plan and repeat the retune process four times.
Hence, the robotic system performed $8$ trials for each object.
The relevant objects as extracted by the LLM from the scene description are \texttt{crumpled paper ball 1}, \texttt{crumpled paper ball 2}, \texttt{whole apple}, \texttt{half-eaten apple}, \texttt{empty glass 1}, \texttt{glass with yellowish liquid}, \texttt{large red trash can}, and \texttt{large white sink} (kitchen sink).

\begin{figure*}[htp!]
    \centering
    \includegraphics[width=\linewidth,trim={0 0 0 0},clip]{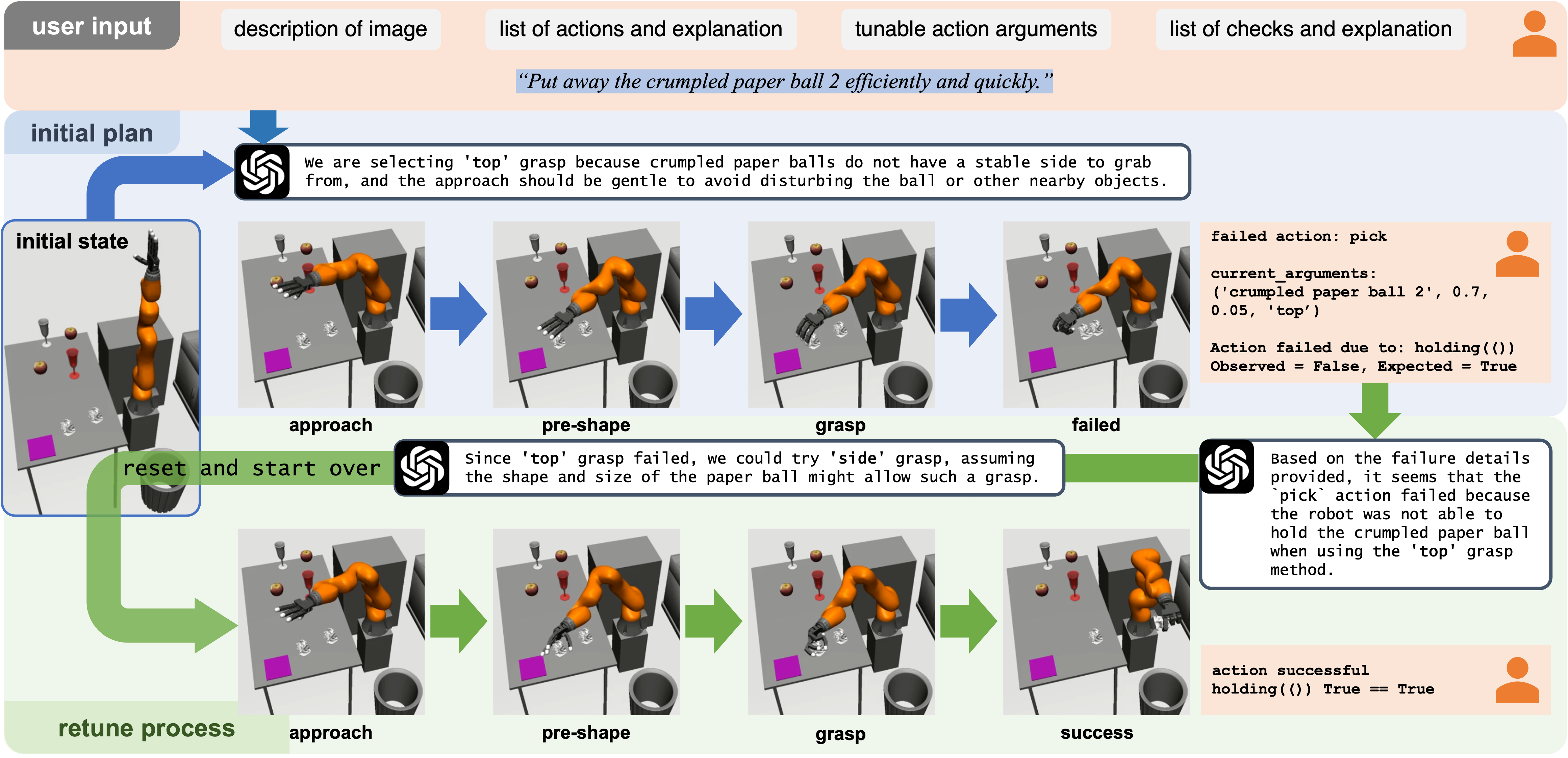}
    \caption{Illustration of the motion re-tune process and the associated LLM's reasoning process using the task of clearing a paper ball. The execution result in the simulation is used as feedback to the LLM for retuning}
    \label{fig:retune_process}
\end{figure*}

Fig.~\ref{fig:retune_process} takes the plan of clearing the further paperball as an example to explain in detail the reasoning process of LLM behind motion retuning.
The tuned parameter is grasp type (see Table~\ref{tab:parameters}).
The LLM started with an initial guess of using \texttt{top} to approach and grasp the paper ball, which is a reasonable choice.
However, the action failed, as the paper ball slipped out of the hand when it closed. Then, based on the output information, the LLM tried a different action parameter and instructed the robot to grasp the paper ball from the side, leading to a successful execution.

\begin{table}[]
\centering
\begin{tabular}
{l @{\hspace{8pt}}l @{\hspace{8pt}}l @{\hspace{8pt}}l @{\hspace{8pt}}l}
\hline
Object                  & $T_{t}$ & $T_{e}$ & $T_{s}$ & SR \\ \hline
\texttt{apple}                   & 18.89$\pm$8.77          & 6.96$\pm$2.98 & 88.20$\pm$1.63 & 87.50 \\ 
\texttt{eaten apple} & 19.75$\pm$6.16 & 4.16$\pm$0.81 & 174.71$\pm$1.43 & 62.50 \\
\texttt{glass empty} & 19.21$\pm$13.51 & 10.47$\pm$4.41 & 190.49$\pm$4.47 & 87.50 \\ 
\texttt{glass full}  & 19.01$\pm$6.12 & 12.78$\pm$9.18 & 180.66$\pm$36.98 & 100.00 \\ 
\texttt{paperball 1}             & 18.70$\pm$7.80 & 9.95$\pm$3.80 & 101.32$\pm$3.75 & 100 \\ 
\texttt{paperball 2}             & 19.16$\pm$5.07 & 13.90$\pm$7.39 & 113.79$\pm$4.83 & 50.00 \\ \hline
\end{tabular}
\caption{Performance metrics: LLM response time (mean$\pm$std) for generating task plan ($T_{t}$, sec), evaluation plan ($T_{e}$, sec), total simulation time ($T_{s}$, sec), and success rate (SR, $\%$). To discriminate the two paper balls, we use ``paperball 1'' to indicate the closer one to the robot base}
\label{tab:motion_tuning_time}
\end{table}

\begin{figure*}[htp!]
    \centering
    \includegraphics[width=\linewidth,trim={0 0 0 0},clip]{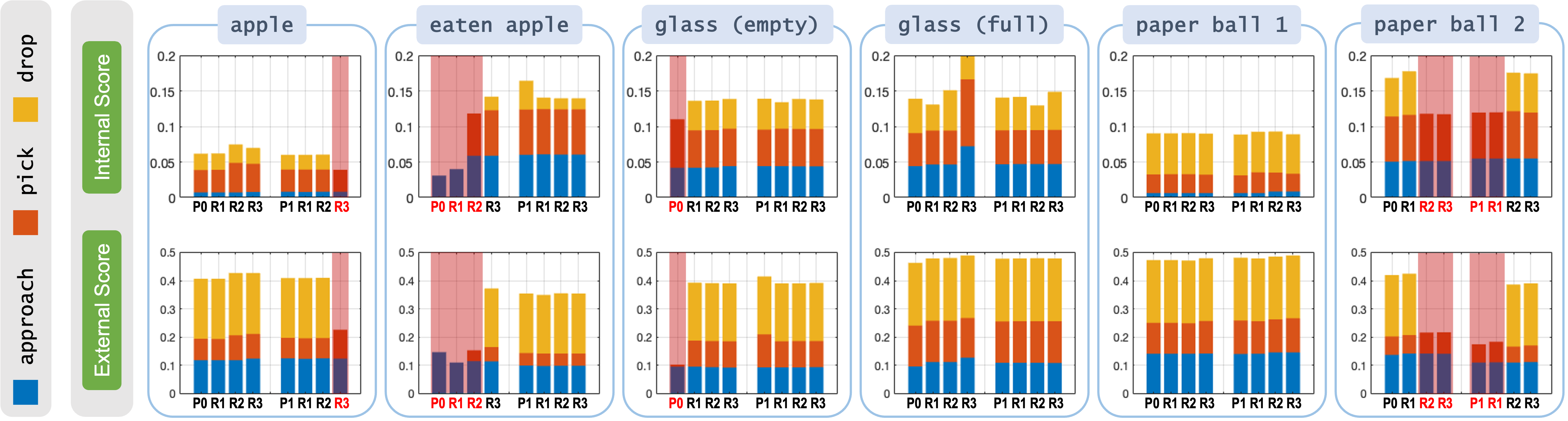}
    \caption{Motion scores after retuning and replanning. P0 and P1 indicate the original task plan and first re-plan, and R1, R2, and R3 indicate the retune of motions in each task plan, respectively. The shaded area in red color indicates failures}
    \label{fig:single_plan_score}
\end{figure*}

Following our proposed framework, LLM generated a successful motion plan for the given task in most cases (see Table~\ref{tab:motion_tuning_time}) with an overall success rate of $81.25\%$.
Most failures can be recovered after 1-2 steps of re-tuning (for \texttt{eaten apple}, \texttt{empty glass}, and second plan of \texttt{paper ball 2}).
When the task is already successful, with retuning, the LLM managed to improve the overall scores of task trials in most plans (see Fig.~\ref{fig:single_plan_score}), indicating an improvement in trajectory for avoiding internal singularities and external obstacles.
It is worth noting that for \texttt{apple} and \texttt{paper ball 2} (first plan), failure occurred after retuning. This is likely because the LLM emphasized motion quality too much without knowing the parameters' boundary for guaranteeing success.

\subsection{Evaluation of Action Contextualization}\label{subsec:exp_contextualization}

\begin{figure*}[htp!]
    \centering
    \includegraphics[width=\linewidth,trim={0 0 0 0},clip]{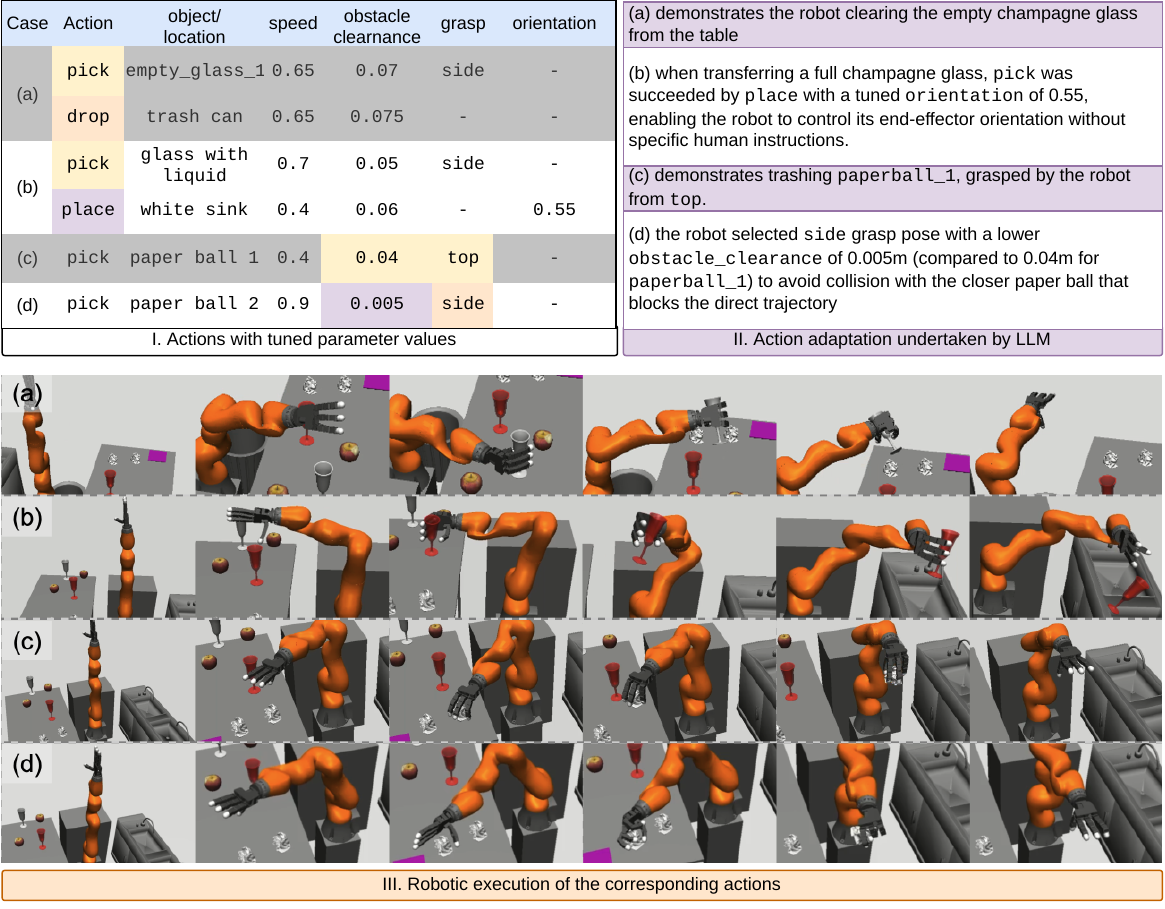}
    \caption{Four comparative experiments ((a) through (d)) to assess action contextualization achieved by the LLM: (I) presents the actions and their tuned parameters from the respective task plans; (II) presents a description of these adaptations performed by the LLM; (III) presents the corresponding simulated robot behaviors}
    \label{fig:motion_contextualization}
\end{figure*}

We followed the same experimental procedure described in the previous section to conduct two comparative experiments.
Fig.~\ref{fig:motion_contextualization} illustrates the robot motions for different objects, namely, the different contexts of the task.
The key actions and their parameters have been presented in Fig.~\ref{fig:motion_contextualization}.
Notice that the LLM was agnostic to each case's specific task constraints and expectations.
Fig.~\ref{fig:motion_contextualization}(a) demonstrated the clearing of an empty champagne glass, where the robot grasped the glass from its \texttt{side} during \texttt{pick}, which was followed by \texttt{drop} into the trash can.
When transferring a full champagne glass in Fig.~\ref{fig:motion_contextualization}(b), \texttt{pick} was succeeded by \texttt{place} with a tuned value of 0.55 for \texttt{orientation}, which enabled the robot to control its end-effector orientation without receiving specific instructions from the human experimenter. The constraint of maintaining the end-effector straight was achieved by LLM tuning the orientation parameter of the action.
In comparison to Fig.~\ref{fig:motion_contextualization}(a), Fig.~\ref{fig:motion_contextualization}(c) illustrates the process of trashing \texttt{paperball 1}, which was grasped by the robot from \texttt{top}.
However, when trashing the relatively farther paper ball 2, as the direct trajectory to the target object is blocked (by the closer paper ball) as Fig.~\ref{fig:motion_contextualization}(d) shows, the robot selected \texttt{side} grasp pose to avoid collision with a much lower \texttt{obstacle\_clearance} of 0.005m, in contrast with 0.04m for paper ball 1.
These cases, once again, demonstrate context-aware planning but for similar objects. The context in these cases is established by the automated feedback prompts received by the LLM via the checks in \texttt{evaluation\_plan}.

These experimental scenarios demonstrate the framework's ability to ``wisely'' choose parameters based on the task's context, benefiting from the human experience inherent in the LLM and performing context-aware planning, leveraging the automated feedback received via \texttt{evaluation\_plan} checks.
A complete example of conversation is available in Appendix.~\ref{appendix:conversation}.

\subsection{Evaluation of Sequential Task Planning}\label{subsec:exp_sce_2}
\begin{figure*}[htp!]
    \centering
    \includegraphics[trim=0cm 0 0 0,clip,width=\linewidth]{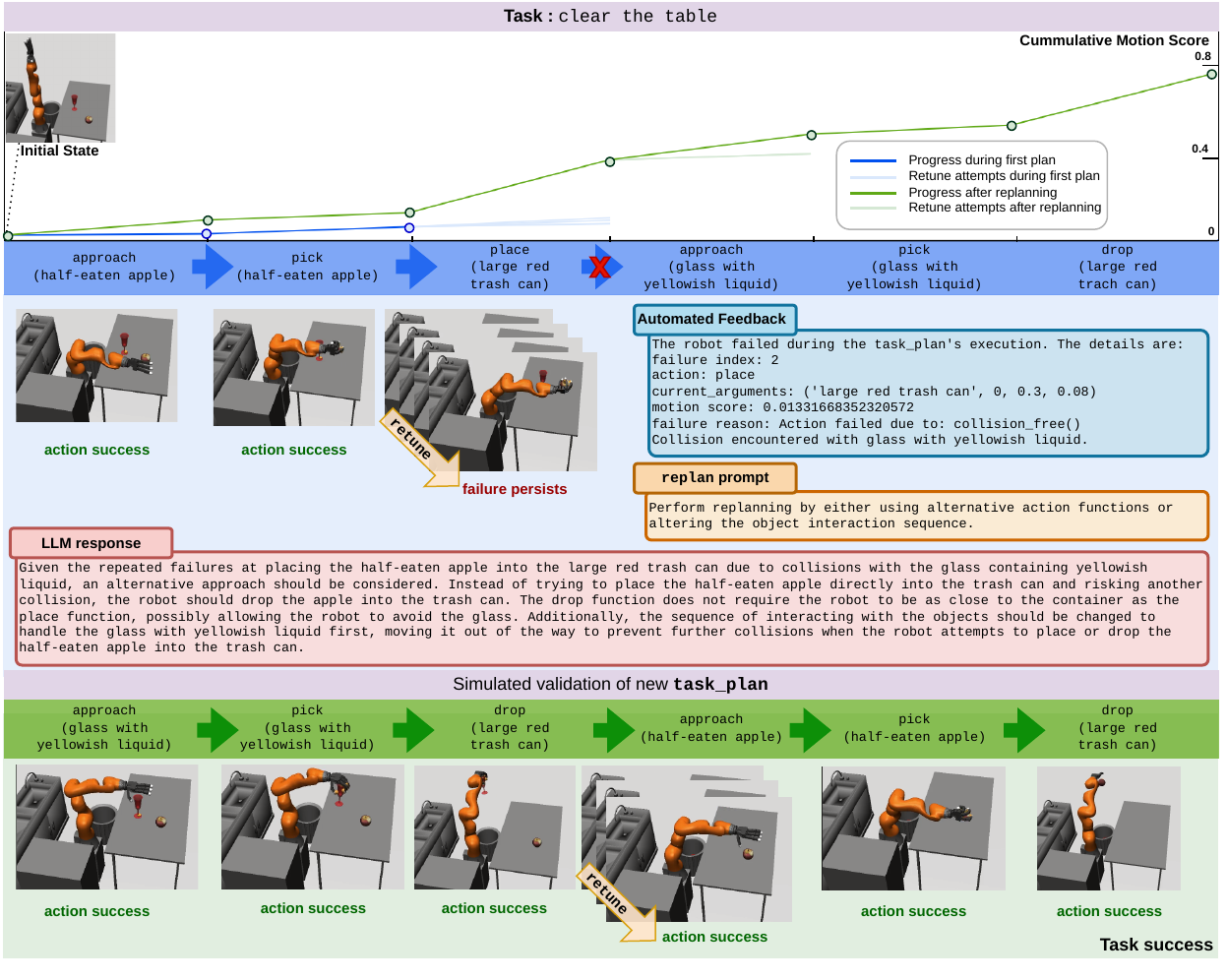}
    \caption{Planning, parameter tuning, and replanning for a simulated instance of the table clearing task}
    \label{fig:experiment2_outline}
\end{figure*}

Next, we investigate planning and motion adaptation in a multi-object manipulation task. The robotic agent is instructed to ``\textit{clear the table}''.
Two objects, \texttt{half-eaten apple} and \texttt{glass with yellowish liquid}, are placed on the table. 
As part of the specified task, the robot is expected to retrieve and put these objects away.
Having formulated a preliminary \texttt{task\_plan} composed of actions from Table~\ref{tab:actions} and \texttt{evaluation\_plan} composed of checks from Table~\ref{tab:checks}, the objective is to tune the action parameters and (or) perform replanning to generate a revised \texttt{task\_plan}, following the feedback from \texttt{evaluation\_plan}.

The evolution of this task is depicted in Fig.~\ref{fig:experiment2_outline}.
For the experiment, we set a limit of 3 \texttt{retune} attempts per action. 
If failure persists, the \texttt{replan} procedure is called. 
The initial \texttt{task\_plan} formulated by the LLM begins with performing the \texttt{approach(half-eaten apple)} action, which is succeeded by \texttt{pick(half-eaten apple)} and then \texttt{place(half-eaten apple)}. The \texttt{place} action fails consecutively and exhausts the three permitted \texttt{retune} attempts. The failure is repeated collision with filled glass, reflected by the \texttt{collision\_free()} check. This constitutes the automated feedback, which, in conjunction with the \texttt{replan} prompt (Fig.~\ref{fig:experiment2_outline}), instructs the LLM to undertake replanning and generates a new \texttt{task\_plan}.
In an attempt to mitigate the repeated collisions with the champagne glass, our framework generates the action to put away the glass before proceeding to the apple. After replanning, the execution of the \texttt{task\_plan} proceeds smoothly, with parameter tuning being required in an action of the new task plan.

\begin{figure}[htp!]
    \centering
    \includegraphics[trim=0cm 0 0.5cm 0,clip,width=\linewidth]{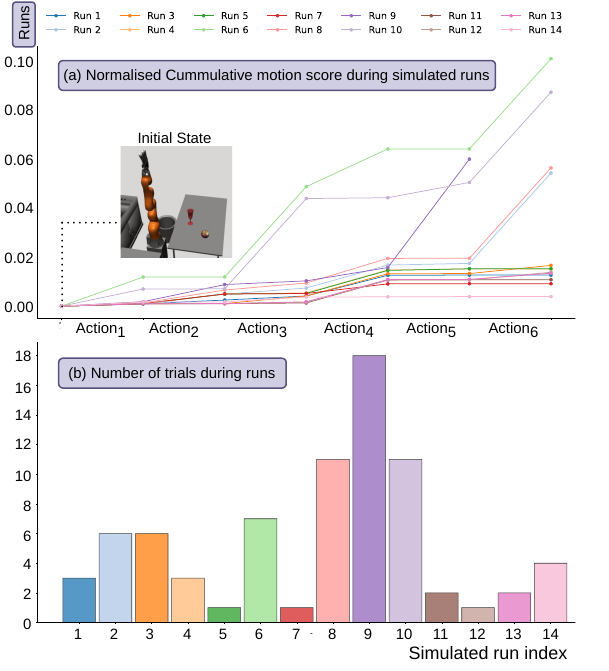}
    \caption{Evaluation of the $14$ simulated instances of the table cleaning task: (a) normalized cumulative scores for the generated task plans, (b) number of trials during simulated validation of the task plans, including both retuning and replanning}
    \label{fig:scoresVsRuns}
\end{figure}
As stated earlier, multiple solutions may exist to accomplish the aforementioned task. The multiplicity in solutions can arise from various sources, including:
\begin{itemize}
    \item Sequence of object-interaction (e.g., first interacting with an eaten apple instead of a filled glass;
    \item Choice of actions (e.g. using \texttt{place} instead of \texttt{drop} while putting the objects away);
    \item Choice of tune-able action parameters (e.g., acting with a high speed as opposed to a slow speed).
\end{itemize}

To find an optimal plan among some of its possible variants, as part of the Planning (Sec. \ref{subsec:planning}) and Adaptation (Sec. \ref{subsec:tuning}) framework, we make $14$ simulated runs of the task with identical environments. 
The sixth run and its evolution have been presented in Fig.~\ref{fig:experiment2_outline}. 
Each run corresponds to an independently generated and tuned \texttt{task\_plan}, which has been scored and validated in simulation. 
As the number of actions may vary across the different task plans, the cumulative score for each plan is normalized by dividing it by the corresponding number of steps in that plan. The evolution of these scores for the different runs is presented in Fig.~\ref{fig:scoresVsRuns}(a). The total number of trial runs for each plan - due to retuning and replanning - is presented in Fig.~\ref{fig:scoresVsRuns}(b).

To proceed with robotic execution, we select the plan with the highest score among the simulated runs. In the current case, this corresponds to the sixth run.

\subsection{Comparison of Different LLM Models}\label{subsec:LLM_comparison}
\begin{figure*}[htp!]
    \centering
    \includegraphics[width=\linewidth,trim={0 0 0 0}, clip]{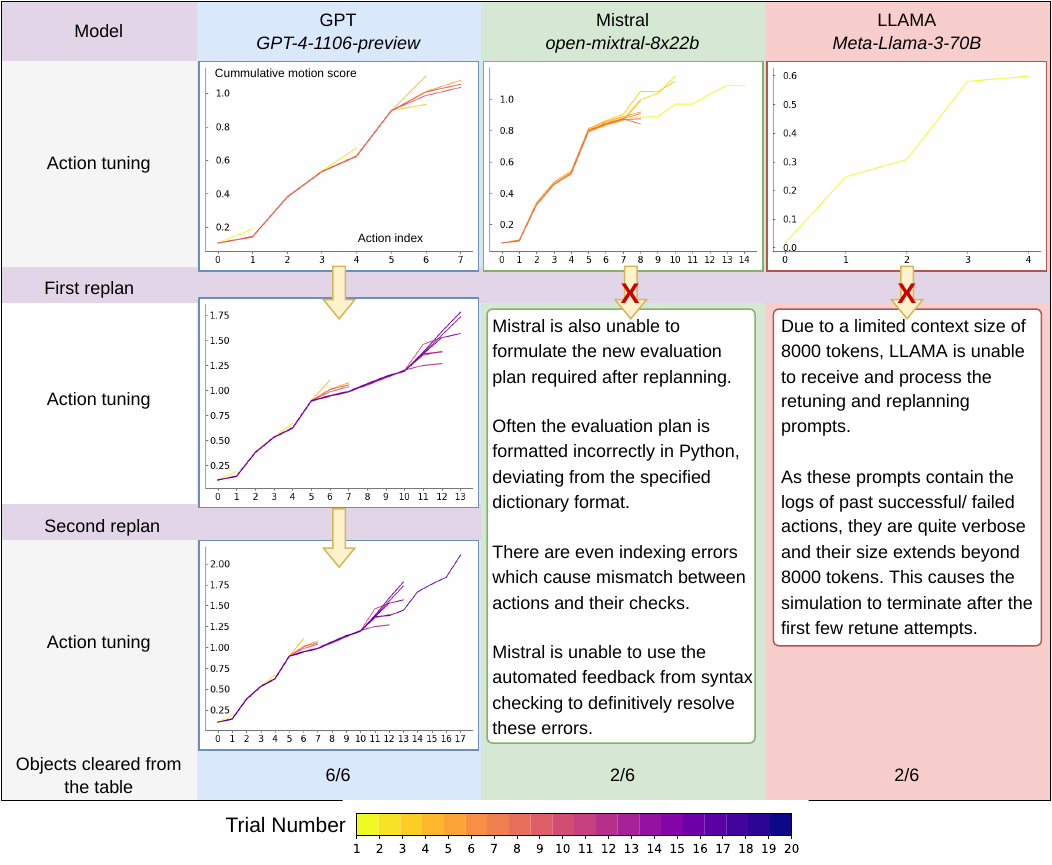}
    \caption{Comparison of performance using different LLMs to perform the table clearing task}
    \label{fig:LLM_comparison}
\end{figure*}

We use the table clearing task as a test to benchmark the planning and action tuning capabilities of different LLM models, chiefly GPT (\texttt{GPT-4-1106-preview}), Mistral (\texttt{open-mixtral-8x22b}) and LLAMA 3 (\texttt{Meta-Llama-3-70B}), with the latter two being open-weight models. The progress of the task is tracked using the cumulative motion score as more actions are performed and the number of objects that were successfully cleared from the table (Fig.~\ref{fig:LLM_comparison}). Using GPT as the LLM for our robotic framework yields the best results.\\
When using Mistral, we observed issues in formulating the evaluation plan. The Python syntax for the generated plan was often incorrect (Appendix.~\ref{appendix:Mistral_bad_planning} Case (a)). In this case, the formulated evaluation plan did not adhere to the specified list format. We also observed that the checks in the formulated evaluation plan may not have a direct correspondence with the task plan (Appendix.~\ref{appendix:Mistral_bad_planning} Case (b)), where the indices 8 and 11 are repeated, making their correspondence with the relevant action from task plan ambiguous. During the simulation, the generated plans went through automated syntax checks, and feedback on errors (if any) was sent back to the LLM to undertake debugging and follow the previously specified planning instructions more closely. However, the LLM was not guaranteed to succeed.\\
When using LLAMA, we observed a seemingly limited understanding of the available actions to undertake the clearing of the table. While the task and evaluation plans were syntactically correct, exclusively the \texttt{pick-place} or \texttt{pick-drop} routine was used (Appendix.~\ref{appendix:LLAMA_bad_planning} Case (a)). The \texttt{pick} action is designed only to retrieve objects that are close enough and within reach. The lack of use of the \texttt{approach} action, designed specifically to overcome this (Table~\ref{tab:actions}), is detrimental to task success. Furthermore, the reasoning accompanying the actions used in the task plan (Appendix.~\ref{appendix:LLAMA_bad_planning} Case (a)) is also less comprehensive compared to its Mistral and GPT counterparts. While it may be possible that with iterative retuning and (or) replanning, this may be overcome, the current context size of 8000 tokens is not big enough to house the entirety of the retune or replan prompts, as after a few retune attempts this limit is hit. These prompts are fairly verbose and can be handled by the GPT and Mistral models. We use the \texttt{GPT-4-1106-preview} model for the remainder of the work.

\subsection{Evaluation of Robustness in Various Scenarios}\label{subsec:random_experiments}
\begin{figure}[h]
    \centering
    \includegraphics[trim=0.2cm 0.5cm 0.5cm 0.1cm,clip,width=\linewidth]{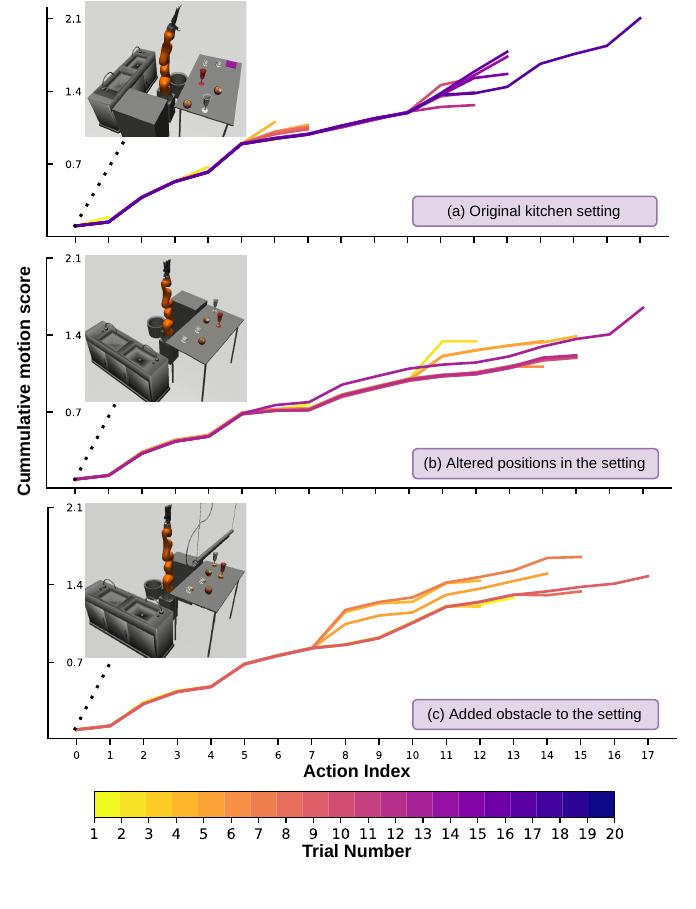}
    \caption{Convergence of motion scores during experiments in various scenarios}
    \label{fig:varied_sims}
\end{figure}

To validate the robustness of our framework in diverse environments, we adjusted the original experimental setup (Fig.~\ref{fig:varied_sims}(a)) and conducted experiments in two new scenarios. 
In the first modified scenario (Fig.~\ref{fig:varied_sims}(b)), there was a randomized alteration in the placement of tabletop items, as well as the kitchen sink, storage shelf, and trash can. 
The second scenario (Fig.~\ref{fig:varied_sims}(c)) introduced an overhead lamp, adding complexity by occluding the robot's operational space and necessitating avoidance of collisions with the lamp. 
Following our framework, the robot completed the table-clearing tasks in both scenarios, as Fig.~\ref{fig:varied_sims} shows, illustrating the motion scores and the convergence process across tuning trials.
For each original and new scenario, the action adaptation was carried out in cycles of 8 trials, consisting of 7 \texttt{retune} attempts and a \texttt{replan} attempt.
It is clear from the figure that the initial plans did not succeed in all these scenarios; replanning was performed twice for each scenario, with the overall retune attempts varying. Throughout the simulation, pursuing action adaptation, the robot was able to complete the task despite the complexity of the scenario.

\subsection{Evaluation on Real Robotic System}\label{subsec:exp_sce_3}

\begin{figure*}
    \centering
    \includegraphics[width=\linewidth]{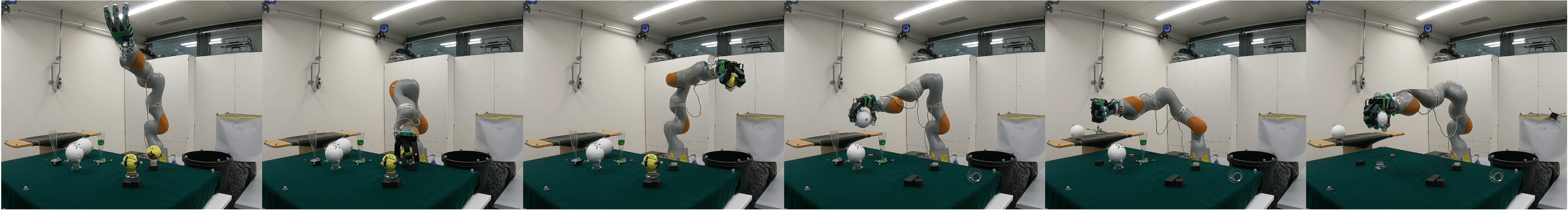}
    \caption{Screenshots of the real robotic arm-hand system performing the sequential task following the retuned task plan}
    \label{fig:real-robot}
\end{figure*}

\begin{figure}
    \centering
    \includegraphics[width=\linewidth]{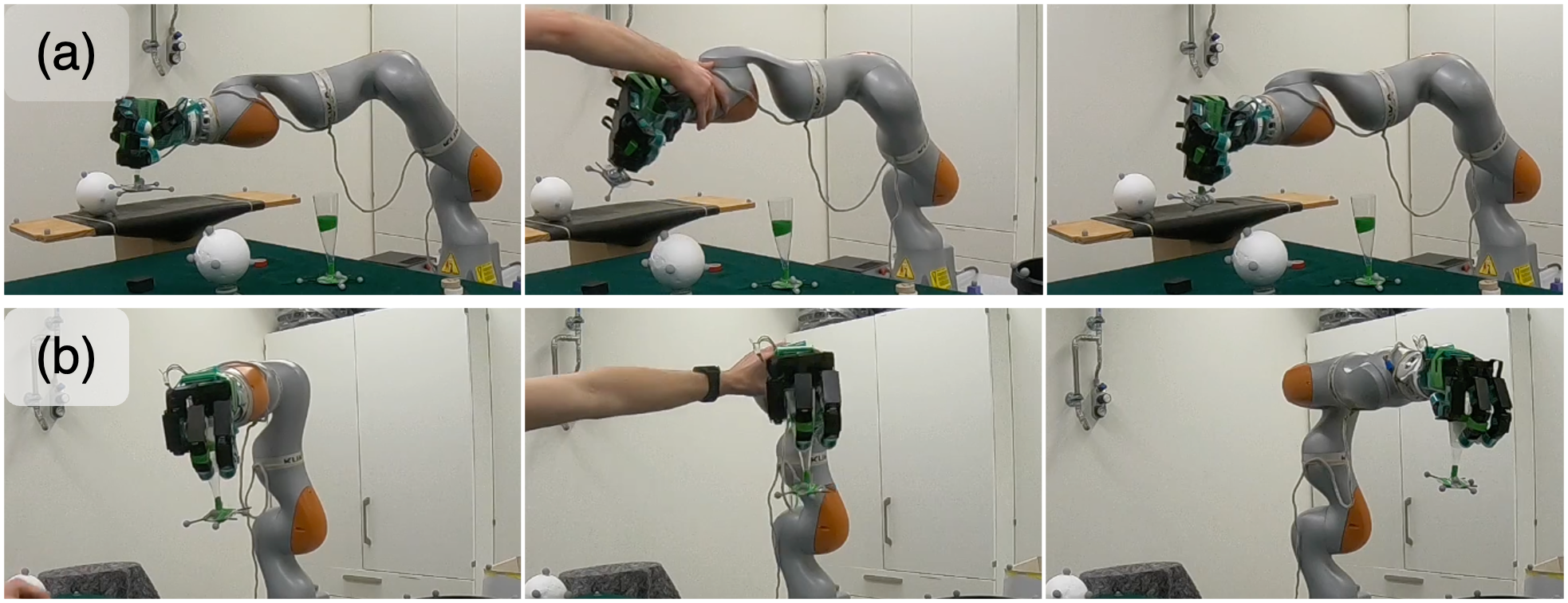}
    \caption{A demonstration of robustness against external disturbance during performing (a) a \texttt{side} grasp and (b) a \texttt{top} grasp, enabled by our DS-based robot controller}
    \label{fig:robust}
\end{figure}

After refining and confirming the task plan, we apply it to control the real robot in completing the set sequential task, as depicted in Fig.~\ref{fig:real-robot}. Additionally, we demonstrate the system's ability to withstand external disruptions through experiments involving intentional interruptions by a human, shown in Fig.~\ref{fig:robust}. The combination of our DS-based controller and the neural distance function for collision avoidance enables the robot to maintain its task orientation, recover from disturbances, and successfully achieve its objective.

\section{Discussion}
\subsection{LLM for Robotic TAMP}
\modify{Our experiments showed that as part of the framework, providing action descriptions (Table~\ref{tab:actions}) to the LLM, in conjunction with the task and scene descriptions, enables it to establish correspondence between the action's tunable parameters and success metrics established by the task (and context in the scene).}
We demonstrated parameter tuning and motion scoring for single-object manipulation tasks with different objects in Sec.~\ref{subsec:exp_sce_1}.
The framework can leverage the human-like intelligence and common sense embedded in the LLM, when tuning action parameters throughout multiple \texttt{retune} and \texttt{replan} procedures (Fig.~\ref{fig:motion_contextualization}).

This was reflected in the gradually improving motion scores (Fig.~\ref{fig:single_plan_score}). \modify{It is also important to note the maximum permissible size of the context that can be accommodated by the LLM model for long-horizon planning and (or) tuning. For instance, the Meta-Llama-3-70B model, with a (currently) limited context size of $8000$ tokens, is unable to perform retuning beyond a few steps as it requires the prior history of successful and failed attempts. On the other hand, \texttt{open-mixtral-8x22b} and \texttt{GPT-4-1106-preview} have context sizes of up to $64000$ and $128000$, respectively.}

When performing robotic manipulation in a complex human environment such as a kitchen, it is often challenging to quantify the optimality of a planned strategy. This complexity can arise due to various obstacles, objects (interactive and non-interactive), and the implicit physical constraints they impose. The ability to leverage contextual information about the task (and environment) from natural language and utilize common sense allows the formulated strategies to avoid failing such constraints. For instance, when performing \texttt{pick} and \texttt{place} actions with the full glass, it is crucial to maintain the orientation of the glass to avoid spillage of its contents. 
While performing similar actions on the half-eaten apple, manually defining an objective function that maximizes execution speed may yield ideal results. 
However, the same objective function may yield undesirable results with the filled glass on account of not prioritizing object orientation over execution speed.
Such task contextualization enables motion adaptation during planning (Fig.~\ref{fig:motion_contextualization}, (a) and (b)) as well as parameter tuning (Fig.~\ref{fig:motion_contextualization}, (c) and (d)) resulting in a robust task execution.

This is also observed in Sec.~\ref{subsec:exp_sce_2}, where a multi-object manipulation task is specified. For example, in Fig.~\ref{fig:experiment2_outline}, the LLM can reason about its persistent failure to \texttt{pick} the eaten apple and formulates a new plan first to put away the filled glass.
\modify{This is interesting as no explicit spatial information about the objects or the robot's whereabouts is provided within the framework.} 
\modify{Using the context, automated feedback from the framework, and following it up with commonsense reasoning, the LLM successfully plans and tunes the \texttt{task\_plan}.}
Generating, simulating, and scoring multiple such plans allows us to pick the highest-scoring plan for robotic execution (Fig.~\ref{fig:scoresVsRuns}).

Essentially, the action tuning and replanning ability can be considered parameter fitting using LLM.
The benefit lies in leveraging human life experience (i.e., prior knowledge) when planning motions for everyday tasks that heavily rely on human life experience and have also been intensively demonstrated by humans.
In comparison to model-based approaches such as optimal control or constrained optimization, LLM-based planning demonstrated the ability to have a suitable initial guess of the motion characteristics without explicit prior knowledge provided by humans, making it possible to embed context-sensitive task parameters, achieving context-aware TAMP autonomously.
In contrast to learning-based approaches, LLM excels in computational efficiency, as it can achieve a satisfactory performance within only a few steps, with no need for collecting large amounts of training data.
These features of LLM can be especially beneficial in dealing with high-dimensionality problems, such as planning long sequential motions and achieving human-like behavior in everyday life.

\subsection{Treading with Caution}
It is worth noting that despite the successful examples in our experiments, LLM also failed in certain cases and showed inconsistency and unpredictability in the performance.

\paragraph{Peculiar behaviors by LLMs}
We observed sporadic peculiarities manifesting logical incoherence during plan generation when we developed our framework. Furthermore, the behavioral unpredictability due to the stochastic nature of LLMs needs to be treated with caution.
The outlier cases facilitated the progressive refinement of our prompts to detect and contain planning irregularities. Such cases typically exhibit missing context, creative but unrealistic solutions, using incorrect object labels, and planning errors due to repeated, redundant, or semantically disoriented actions.
Appendix~\ref{appendix:peculiar_planning} demonstrates a few peculiar instances. Case (a) demonstrates repeated instructions. This may cause a failure to be registered while executing the extra \texttt{place} while the task has already been completed.
Case (b) demonstrates LLM having somewhat incorrectly understood the \texttt{approach} action. The \texttt{approach} action is to get close enough to an object to grasp it. However, approaching the trash can with the intent to grasp it is different from hovering on top of it to then \texttt{drop} the apple.
Case (c) shows the LLM oddly \texttt{pick}ing the apple only to \texttt{place} it back on the table. This could be attributed to the missing context. From the scene description, the LLM may have failed to infer that the apple was already on the table.
Case (d) follows an instance after \texttt{replan}. As evidenced by the accompanying reasoning, we can say that the LLM assumed that through \texttt{approach}, the robot gets close enough to the paper ball to manipulate it indirectly (e.g., nudging or pushing) so that \texttt{drop} will next result in the ball falling into the trash can. It is also interesting to note that this change was due to purposefully skipping the original \texttt{pick} due to its repeated failure. The failure from frustration seems to have forced it to push its bounds on creativity.
Generating the \texttt{task\_plan} and \texttt{evaluation\_plan} largely identifies and eliminates such cases as they fail. These plans are then replaced by \texttt{replan} or a more optimally scoring counterpart generated in a different simulated run. 

\paragraph{Error detection and resolution}
The aforementioned instances demonstrate how the LLM is susceptible to erroneous planning. A method to detect and remove such planning inconsistencies is using \emph{post hoc} verification: checking the status of each motion and retuning/replanning the motions in simulation before executing on a real robotic system, e.g., using the proposed \texttt{evaluation\_plan}.

First, plans that violate the logic will be detected due to a mismatch between observations and desired values from self-placed checks.
Second, the proposed framework conducts simulation-based validation to assess the feasibility of the robotic system executing the planned actions. Even if an erroneous plan manages to pass the placed evaluation checks, it will fail in the simulation.
Lastly, the generated plan has a Python format, and code compilation prior to execution serves as a tertiary check to identify indexing errors or ambiguous correspondence between actions and checks. When a plan is flagged, the LLM is prompted to retune, replan, or correct syntax. 
By generating multiple independent plans following the same pipeline and scoring them independently, the incorrect or sub-optimal plans get dropped out. The framework detects and (or) resolves such instances as follows:
\begin{itemize}
    \item The plan contains repeated actions: Such plans are flagged during simulation-based validation, as it is impossible to achieve success when repeating an action with the same parameters but via different starting conditions;
    \item Semantically disoriented actions: When an action is used incorrectly (e.g., using place instead of drop), the plan once again gets flagged either by the self-placed action-specific predicate checks or during simulation due to physical constraints. This happens because the behavior of the robot controller responsible for executing the specified action does not match up with the expected outcome (either in simulation or during self-evaluation);
    \item Performing redundant actions: At times, the plan included redundant actions, such as picking an object from the table only to place it back on the table. Such actions are non-detrimental to the task's success or simply get flagged during simulation due to code equivalence.
\end{itemize}

\paragraph{Excessive retuning is risky} The key objective of motion adaptation (\texttt{retune}) in our framework is to achieve task success and improve motion quality, not necessarily achieve optimality, and must be used as such.
The parameter tuning step aims to let LLM robustly assess the feasibility of the previously generated \texttt{task\_plan}.
While we score these motions, they serve more as hindsight to LLM.
As stated before, \texttt{retune} lacks an objective function (or even a metric) to quantify the task performance to arrive at a set of globally optimum parameters.
Forcing \texttt{retune} post achieving success did not necessarily exhibit a positive correlation with the accompanying motion scores (see Fig.~\ref{fig:single_plan_score}).
\modify{This might indicate a limitation in LLM: although the physical meaning of each metric is described, the LLM may struggle to adjust the robot system without a detailed model.
This may imply a disconnect between the LLM's abstract and high-level task understanding and the specific, real-world requirements of robotic operations, hence highlighting the importance of integrating detailed robotic models with LLMs for more challenging action tuning.}
Thus, parameter tuning should be used only to achieve task success.

\paragraph{The performance fluctuation in LLM} to use LLM for TAMP, it is crucial to ensure stable performance by the LLM. However, we have noticed that the response quality and response time of LLM can vary greatly depending on the time of the day. The best performance is typically achieved during $0800-1600$ CET; degraded performance is outside this time window. While not encountered recently, \textit{lazy planning} is also observed in LLM, especially for multi-object manipulation tasks. When this occurs, the LLM usually prescribes a \texttt{pick}-\texttt{place} routine for a few objects and then comments in \texttt{task\_plan} to carry on similarly, without providing relevant actions and parameters.
The retuning and replanning steps in our framework can only help prevent such plans from being used to instruct the real robot but cannot eliminate this issue.

\subsection{Integrating LLM into Real Robotic System}
\modify{The experimental results demonstrate that the proposed framework is able to leverage LLMs into creating context-aware motions, enhancing the robot's adaptability to underlying task constraints and requirements.}
The main hurdle when integrating LLMs with actual robotic systems is the real-world interaction, mainly due to the LLM's inability to adjust robot movements in real-time, such as responding to disturbances or avoiding collision in a changing environment.
Therefore, it’s imperative for the controller to equip the robot with robust and adaptive motions while avoiding both self-collision and collisions with the external environment in real time.
We first tested this in simulation by setting up new scenarios for task execution. Between the original and the first new scenario (Figs.~\ref{fig:varied_sims} (a) and (b)), adding randomness to the positions of objects in the scene allowed us to test the adaptability of the controllers to the action-parameter combinations.
In the second setting (Fig.~\ref{fig:varied_sims}(c)), adding a lamp as an additional obstacle in the workspace allowed us to test collision avoidance.

\modify{In our real-world tests, the DS-based controller significantly improved the robot's ability to adjust and maintain stable performance, even under deliberate human interference. Additionally, the neural distance function facilitated immediate collision detection by inferring the robot's collision status in real time.
This avoids the voxelization of the real world \cite{huang2023voxposer}, which can hardly be applied to deal with a dynamic environment in real time. 
This highlights the essential role of an adaptive controller and collision detection algorithms in ensuring successful task execution by robotic systems when integrating LLMs into real robotic applications, underlining its importance for seamless motion execution.}

\section{Conclusion}
In this paper, we presented an LLM-based framework for robotic TAMP that incorporates \emph{action contextualization}, adapting robot motion to the task context using insights from human experience embedded in the LLM.
Our approach uniquely allows for online adjustment and replanning based on simulated outcomes, mitigated by motion scores, to address the LLM's lack of awareness of robotic models and dynamics, ensuring more appropriate motion profiles for actual robots.
The integration of these elements offers a robust solution bridging task planning and robotic execution.
Notably, the success of our real robotic experiments is largely attributed to the dynamical system-based robot controller \modify{and collision avoidance ability}, which facilitates adaptive and reactive robot behavior in real time.
\modify{The proposed framework can easily be extended to other real-world manipulation tasks, where context is crucial during planning and execution.}
\modify{The framework facilitates a straightforward integration with modular control strategies, such as Dynamic Movement Primitives. Essential to this integration is an appropriate parameterization of motion elements, which should capture the contextual details of the task, thus leveraging the potential benefits of LLM.} \modify{Furthermore, by defining new actions and adding their descriptions to the existing stack, it is also possible to use user-defined robotic behavior during task execution.}
\modify{Similarly, defining new checks and adding their descriptions enables verification of the task execution.}
\modify{These aspects collectively contribute to an easy-to-program robotic agent for real-world task execution.}
Despite the successes, the LLM's \emph{black-box} nature presents inherent limitations and risks, warranting caution.
In our future work, we plan to explore more complex interactions with the environment, validating our approach in unknown environments, including scenarios such as exploration with mobile manipulators or humanoid robots. Additionally, we will explore selective score prioritization for motion refinement and the development of a local skill model within the LLM for enhanced task-specific learning.
\modify{Lastly, we aim to generalize the formulation of our action sets to facilitate seamless integration with modular control approaches, significantly expanding the applicability of our framework.}


\ifCLASSOPTIONcaptionsoff
  \newpage
\fi

\printbibliography

\clearpage
\newpage
\onecolumn

\appendices
\section{Checks}
\label{app:checks}
\floatname{algorithm}{Check}
\begin{algorithm}
\caption{\texttt{can\_grasp(object\_to\_grasp)}}\label{alg:can_grasp}
\begin{algorithmic}
\ENSURE Returns \texttt{True} if the robot can grasp the object, otherwise \texttt{False}
\STATE $distance \gets$ $\|object\_to\_grasp_{pos} - gripper_{pos} \|_2$
\IF{$distance < obstacle\_clearance$}
    \RETURN \texttt{True}
\ELSE
    \RETURN \texttt{False}
\ENDIF
\end{algorithmic}
\end{algorithm}
\vspace{-0.5\baselineskip}

\begin{algorithm}
\caption{\texttt{holding()}}\label{alg:holding}
\begin{algorithmic}
\ENSURE Returns \texttt{True} if the robot is holding an object, otherwise \texttt{False}
\STATE $num\_contact\_points \gets$ Number of contacts registered by the gripper's fingers
\IF{$num\_contact\_points < 2$}
    \RETURN \texttt{False}
\ELSE
    \RETURN \texttt{True}
\ENDIF
\end{algorithmic}
\end{algorithm}
\vspace{-0.5\baselineskip}

\begin{algorithm}
\caption{\texttt{at\_location(object, location)}}\label{alg:at_location}
\begin{algorithmic}
\ENSURE Returns \texttt{True} if the \texttt{object} is at the specified \texttt{location}, otherwise \texttt{False}
\STATE $distance \gets$ minimum distance between $object$ and $location$ meshes
\IF{$distance < 10cm$}
    \RETURN \texttt{True}
\ELSE
    \RETURN \texttt{False}
\ENDIF
\end{algorithmic}
\end{algorithm}
\vspace{-0.5\baselineskip}

\begin{algorithm}
\caption{\texttt{can\_reach(goal)}}\label{alg:can_reach}
\begin{algorithmic}
\ENSURE Returns \texttt{True} if it is feasible for the robot to reach the goal position from the current state, otherwise \texttt{False}
\STATE $path \gets$ IK\_Solution(start = $gripper_{pos}$, end = $goal_{pos}$)
\IF{exist($path$)}
    \RETURN \texttt{True}
\ELSE
    \RETURN \texttt{False}
\ENDIF
\end{algorithmic}
\end{algorithm}
\vspace{-0.5\baselineskip}

\begin{algorithm}[hp!]
\caption{\texttt{check\_motion\_health()}}\label{alg:check_motion_health}
\begin{algorithmic}
\ENSURE Returns \texttt{True} if the robot’s motion during the preceding action was safe for its hardware, otherwise \texttt{False}
\IF{$motion\_score > 0$ AND is\_finite($motion\_score$)}
    \RETURN \texttt{True}
\ELSE
    \RETURN \texttt{False}
\ENDIF
\end{algorithmic}
\end{algorithm}
\vspace{-0.5\baselineskip}

\section{Planning issues when using Mistral}\label{appendix:Mistral_bad_planning}
\begin{reply}[Case (a)]
\vspace{-0.5\baselineskip}
\begin{lstlisting}
evaluation_plan = [
    (0, {'can_grasp': ('crumpled paper ball 1', 'top')}, (True,)),
    (0, {'collision_free': ()}, ('',)),
    (0, {'timeout': ()}, (True,)),
    (0, {'check_motion_health': ()}, (True,)),
    (1, {'holding': ()}, (True,)),
    (1, {'collision_free': ()}, ('',)),
    (1, {'timeout': ()}, (True,)),
    (1, {'check_motion_health': ()}, (True,)),
    ....
    (16, {'holding': ()}, (True,)),
    (16, {'collision_free': ()}, ('',)),
    (16, {'timeout': ()}, (True,)),
    (16, {'check_motion_health': ()}, (True,)),
    (17, {'at_location': ('glass with yellowish liquid', 'large white sink')}, (True,)),
    (17, {'collision_free': ()}, ('',)),
    (17, {'timeout': ()}, (True,)),
    (17, {'check_motion_health': ()}, (True,))
]
\end{lstlisting}
\vspace{-0.5\baselineskip}
\end{reply}

\begin{reply}[Case (b)]
\vspace{-0.5\baselineskip}
\begin{lstlisting}
evaluation_plan = [
    (0, {'can_grasp': ('crumpled paper ball 1', 'top'), 'collision_free': (), 
    'timeout': (), 'check_motion_health': ()}, (True, '', True, True)),
    (1, {'holding': (), 'collision_free': (), 
    'timeout': (), 'check_motion_health': ()}, (True, '', True, True)),
    ...
    (8, {'can_reach': ('storage shelf', 'side'), 'collision_free': (), 'timeout': (),
    'check_motion_health': ()}, (True, '', True, True)),
    (8, {'at_location': ('whole apple', 'storage shelf'), 'collision_free': (), 
    'timeout': (), 'check_motion_health': ()}, (True, '', True, True)),}
    (9, {'can_grasp': ('half-eaten apple', 'side'), 'collision_free': (), 'timeout': (),
    'check_motion_health': ()}, (True, '', True, True)),
    (10, {'holding': (), 'collision_free': (), 'timeout': (), 'check_motion_health': ()}, 
    (True, '', True, True)),
    (11, {'can_reach': ('storage shelf', 'side'), 'collision_free': (), 'timeout': (),
    'check_motion_health': ()}, (True, '', True, True)),
    (11, {'at_location': ('half-eaten apple', 'storage shelf'), 'collision_free': (), 
    'timeout': (), 'check_motion_health': ()}, (True, '', True, True)),
    ...
]
\end{lstlisting}
\vspace{-0.5\baselineskip}
\end{reply}

\section{Planning issues when using LLAMA}\label{appendix:LLAMA_bad_planning}
\begin{reply}[Case (a)]
\vspace{-0.5\baselineskip}
\begin{lstlisting}
task_plan = [
    (0, "pick", ("crumpled paper ball 1", 0.5, 0.01, "top")),
    # Pick up the first crumpled paper ball from the top with moderate speed and 
    # obstacle clearance.
    (1, "drop", ("large red trash can", 0.5, 0.01)),
    # Drop the crumpled paper ball into the trash can with moderate speed and 
    # obstacle clearance.
    (2, "pick", ("crumpled paper ball 2", 0.5, 0.01, "top")),
    # Pick up the second crumpled paper ball from the top with moderate speed and 
    # obstacle clearance.
    (3, "drop", ("large red trash can", 0.5, 0.01)),
    # Drop the second crumpled paper ball into the trash can with moderate speed 
    # and obstacle clearance.
    (4, "pick", ("whole apple", 0.5, 0.01, "top")),
    # Pick up the whole apple from the top with moderate speed and obstacle clearance.
    (5, "place", ("storage shelf", 0.5, 0.01, 0.5)),
    # Place the whole apple on the storage shelf with moderate speed, obstacle clearance, 
    # and maintaining original orientation.
    (6, "pick", ("half-eaten apple", 0.5, 0.01, "top")),
    # Pick up the half-eaten apple from the top with moderate speed and obstacle clearance.
    (7, "place", ("storage shelf", 0.5, 0.01, 0.5)),
    # Place the half-eaten apple on the storage shelf with moderate speed, obstacle clearance, 
    # and maintaining original orientation.
    (8, "pick", ("empty glass 1", 0.5, 0.01, "side")),
    # Pick up the empty glass from the side with moderate speed and obstacle clearance.
    (9, "place", ("storage shelf", 0.5, 0.01, 0.5)),
    # Place the empty glass on the storage shelf with moderate speed, obstacle clearance,
    # and maintaining original orientation.
    (10, "pick", ("glass with yellowish liquid", 0.5, 0.01, "side")),
    # Pick up the glass with yellowish liquid from the side with moderate speed and 
    # obstacle clearance.
    (11, "place", ("storage shelf", 0.5, 0.01, 0.5)),
    # Place the glass with yellowish liquid on the storage shelf with moderate speed, 
    # obstacle clearance, and maintaining original orientation.
]
\end{lstlisting}
\vspace{-0.5\baselineskip}
\end{reply}

\section{Instance of Peculiar Planning by LLM}\label{appendix:peculiar_planning}

\begin{reply}[Case (a)]
\vspace{-0.5\baselineskip}
\begin{lstlisting}
    task_plan = [
        # Approach the glass with yellowish liquid with moderate speed and a grasp that
        # allows for a secure pick-up while avoiding obstacles.
        (0, 'approach', ('glass with yellowish liquid', 0.7, 0.01, 'side')),
        # Pick the glass from the side to avoid obstructing the grip if the glass is
        # wider at the top. Moderate speed chosen for better accuracy
        (1, 'pick', ('glass with yellowish liquid', 0.7, 0.01, 'side')),
        # Move to the storage shelf, identified as the most logical place for kitchenware,
        # with care to maintain speed that balances efficiency and safety of liquid
        # containment.
        (2, 'place', ('storage shelf', 0.9, 0.7, 0.01)),
        # The place action assumes the robot is at the location with the object, we do not
        # need to use approach beforehand, and must be careful with the orientation of
        # the glass since it contains liquid.
        (3, 'place', ('storage shelf', 0.9, 0.7, 0.01))
    ]
\end{lstlisting}
\vspace{-0.5\baselineskip}
\end{reply}

\begin{reply}[Case (b)]
\vspace{-0.5\baselineskip}
\begin{lstlisting}
task_plan = [
    # Approach the whole apple from the side at moderate speed for better accuracy
    (0, 'approach', ('whole apple', 0.7, 0.02, 'side')),
    # Pick the apple from the side with low speed for control and precision
    (1, 'pick', ('whole apple', 0.3, 0.02, 'side')),
    # Approach the red trash can quickly as precision is less of a concern
    (2, 'approach', ('large red trash can', 0.7, 0.02, 'side')),
    # Drop the apple into the trash can without worrying about orientation
    (3, 'drop', ('large red trash can', 0.7, 0.02))
]
\end{lstlisting}
\vspace{-0.5\baselineskip}
\end{reply}

\begin{reply}[Case (c)]
\vspace{-0.5\baselineskip}
\begin{lstlisting}
task_plan = [
    # Approach the whole apple with moderate speed and precision.
    (0, 'approach', ('whole apple', 0.5, 0.02, 'top')),
    # Moderate speed for precise movement. The clearance is set just tight enough for
    # a good approach, picking from the top to avoid damaging the fruit.
    # Pick up the whole apple once in reach.
    (1, 'pick', ('whole apple', 0.5, 0.02, 'top')),
    # Same speed as approach for consistency; we use top grasp to gently pick up the apple
    # without squeezing it.
    # Move to the white table's location to place the apple down.
    (2, 'place', ('white table', 0.5, 0.5, 0.02)),
    # This orientation ensures the apple is placed on the table without rolling,
    #with moderate speed for efficiency and care.
]
\end{lstlisting}
\vspace{-0.5\baselineskip}
\end{reply}

\begin{reply}[Case (d)]
\vspace{-0.5\baselineskip}
\begin{lstlisting}
task_plan = [
    # The robot approaches the crumpled paper ball at a speed that has successfully worked 
    # before.
    (0, 'approach', ('crumpled paper ball 2', 0.5, 0.01, 'top')),
    # Instead of picking it up, the robot now attempts to nudge the paper ball into the
    # trash can.
    (1, 'drop', ('large red trash can', 0.5, 0.01)),
]
\end{lstlisting}
\vspace{-0.5\baselineskip}
\end{reply}

\onecolumn
\setlength{\parindent}{0pt}
\section{Prompt and Reply}\label{appendix:conversation}

\begin{prompt}[Prompt 1: task plan generation]
The image shows an indoor scene with a white table at the center, scattered with various objects. On the table, there are crumpled paper balls, a whole apple, a half-eaten apple, and two different types of glasses, one of which appears to be empty and one partially filled with a yellowish liquid. To the left, there is a large white sink with a faucet against a tiled wall. A storage shelf is also kept near the table. In the background, there are two doors, one closed and one ajar, and a folded white chair next to the table. On the right side, a large red trash can is visible, filled with discarded items including colorful plastic wrappers. The room has a clinical or institutional feel, possibly a break room or a workshop space.

The list of recognized objects is:

objects = [`white table', `half-eaten apple', `glass with yellowish liquid', `large red trash can']

The list of recognized locations is:

locations = [`large white sink', `faucet', `tiled wall', `closed door', `ajar door', `folded white chair', `clinical room', `storage shelf']

There is a robot, labeled `robot', that can only manipulate ONE object at a time. The robot accepts commands in the form of `action functions' written in Python. These action functions, which can be imported from the `action\_functions' library are:
\\

drop(location: str, speed: float, obstacle\_clearance: float) $->$ None: \# Goes to `location' and drops the grasped object over it. It is not advised to use the approach function directly before this one.
\\

approach(object\_to\_grasp: str, speed: float, obstacle\_clearance: float, grasp: str) $->$ None:  \# Moves the robot close to `object\_to\_grasp' so that the object is in robot`s reach
\\

place(location: str, orientation: float, speed: float, obstacle\_clearance: float) $->$ None: \# Positions the `grasped\_object' on/at/in the `location' and release the grasp. It is not advised to use the approach function directly before this one.
\\

pick(object\_to\_grasp: str, speed: float, obstacle\_clearance: float, grasp: str) $->$ None: \# Instructs the robot to pick up the `object\_to\_grasp', if it is close enough
\\

The `speed' argument for `approach', `pick', `drop', and `place' functions, assumes a value in [0,1] and regulates how fast the robot moves. The closer the the value is to 1 the faster the robot moves. moving at a higher speed is faster but might result in a jerky and less precise motion.
\\

The `orientation' argument for the `place' and `drop' functions, regulates how crucial it is for the robot to maintain the original orientation of the object that the robot is holding. A value closer to 1, instructs the robot to strictly maintain the orientation, but may result in difficulty to avoid external perturbations or obstacles.
\\

The `grasp' argument for `approach' and `pick' assumes one of the two values (`top', `side'), where `top' instructs the robot to approach or pick the object from the top and select `side' instructs the robot to approach or pick the object from the side.
\\

The `obstacle\_clearance' for `drop', `approach', `place', and `pick' functions define how close the robot can get from an object (including the one it is trying to grasp in the pick action) before starting to avoid it. The distance is in meters. Small values allow the robot to get closer to obstacles and usually give a better chance of reaching the object, picking it up, and holding it. Typically values are between 0.005 and 0.05 although values out of this range are possible.
\\

The actions described in these functions are the only motions known to the robot. The task for the robot is: ``put away the `half-eaten apple' and the `glass with yellowish liquid' ''. First, explain how you are going to solve the task, why each step is executed, and how it makes sense to a human that you do it this way. Then, using the actions functions, `objects' and `locations', define a task plan as a Python list of tuples (named `task\_plan'), for the robot to follow. The action\_functions from the `task\_plan' will be directly run on the robot. Each element of the list is a tuple of the form (action number, action function name string, (arguments)). For each action, use object and task-specific arguments.
\\

The first index of the plan should be 0. Take into account that some locations might be outside of the robot's reach and you might only be able to use `drop' to put an object there. Do not make any assumptions. For the task plan, output a single code block. With each action include as the comment the reasoning behind the action and its parameters. Assume the necessary modules containing action functions have already been loaded. Write only a single assignment statement creating the full `task\_plan'. Do not abbreviate anything in the code you are writing. No matter how long it is, write it in full.
\end{prompt}

\begin{reply}[Reply 1]
To accomplish the task, we need to create a plan that follows a sequence of picking and placing actions. The robot needs to identify the objects to be manipulated (half-eaten apple and glass with yellowish liquid), safely pick them up, navigate to an appropriate location to place or drop them, and then perform the placement or dropping action.
\\

The robot can only handle one object at a time, so it will first deal with the half-eaten apple and then the glass with yellowish liquid. Since the only actions the robot can perform are approach, pick, place, and drop, we have to carefully decide the order and the specific parameters for each action to ensure a smooth execution. Also, since some locations might be hard to reach, we will have to use the drop functionality if necessary.
\\

Given these considerations, we can devise the following plan:
\begin{enumerate}
    \item Approach the half-eaten apple, ensuring that we have proper obstacle clearance.
    \item Pick up the half-eaten apple, with an appropriate grasp to securely hold the apple.
    \item Place the half-eaten apple in the large red trash can and release it gently to avoid spills or disturbing the trash can contents.
    \item Approach the glass with yellowish liquid, again ensuring obstacle clearance.
    \item Pick up the glass with the yellowish liquid with a grasp that prevents spilling.
    \item Because the trash can might be hard to reach inside, we use a \texttt{drop} instead of a \texttt{place} to dispose of the glass.
\end{enumerate}
Choosing proper speed and orientation values will help to ensure the robot moves efficiently without sacrificing accuracy or safety. We should choose slower speeds to pick and place/drop since these actions require more precision. We'll assume that the half-eaten apple and the glass can be oriented freely when dropped into the trash can (low orientation strictness) and that a closer obstacle clearance is preferable for a better chance of precise picking and placing.

\begin{lstlisting}
# Define the task plan
task_plan = [
    # Approach the half-eaten apple with moderate speed for better precision
    # and sufficient obstacle clearance
    (0, 'approach', ('half-eaten apple', 0.5, 0.03, 'side')),
    
    # Pick the apple from the side to make sure we have a secure grip before lifting
    (1, 'pick', ('half-eaten apple', 0.5, 0.03, 'side')),
    
    # Place the apple in the large red trash can with less concern for its orientation
    (2, 'place', ('large red trash can', 0.2, 0.5, 0.03)),
    
    # Approach the glass with yellowish liquid slowly to avoid spilling the liquid
    (3, 'approach', ('glass with yellowish liquid', 0.5, 0.05, 'top')),
    
    # Pick the glass from the top to prevent spilling the liquid while maneuvering it
    (4, 'pick', ('glass with yellowish liquid', 0.5, 0.05, 'top')),
    
    # Drop the glass in the large red trash can without needing to maintain its orientation
    (5, 'drop', ('large red trash can', 0.5, 0.05))
]
\end{lstlisting}
\end{reply}

\begin{prompt}[Prompt 2: evaluation plan generation]
The robot may not be able to execute an action function or encounter object collision during execution. Thus, it is required to check for completion of each action function after they have been performed.\\

For this, we define some `checking functions' written in Python. These checking functions, which can be imported from the `checking\_functions' library are:
\\

can\_grasp(object\_to\_grasp: str, grasp: str) $->$ bool: \# Returns True if the robot is close enough to the `object\_to\_grasp' to securely grasp it with the determined grasp `side' or `top'
\\

holding() $->$ bool \# Returns True if the robot is holding an object
\\

at\_location(object: str, location: str) $->$ bool: \# Returns True if the `object' is at the `location'
\\

collision\_free() $->$ str: \# If the robot encounters a collision while executing the preceding action, returns the object label string. Otherwise, `'.
\\

timeout() $->$ Bool: \# Returns True if the preceding action was executed in a timely fashion
\\

check\_motion\_health() $->$ bool: \# Returns True if the robot's motion during the preceding action was safe for its hardware
\\

can\_reach(goal: str, grasp: str) $->$ bool: \# Returns True if it is feasible for the robot to reach the `goal' object or location from the current state from the side determined by the grasp argument `side' or `top'. Objects that are out of the workspace will always return false.
\\

The grasp argument is the same as the one in the `approach' and `pick' functions. It assumes one of the two values {`top', `side'}
\\

Using the `checking\_functions', `locations', and `objects', define an evaluation plan (named `evaluation\_plan') to verify the successful execution of each action. Additionally, for each action verify without fail:
\begin{itemize}
    \item collision-free
    \item timely motion
    \item motion health
\end{itemize}
Output this plan as a Python list of tuples, where each tuple is of the form (action number int, dictionary with `check\_function' names as keys and a tuple of arguments as value, tuple of expected outputs). Do not assume any other object or location, beyond those in `object\_labels'. Each tuple is meant to be checked after the action with the corresponding number. Generate the entire plan. No reasoning, direct output.
\end{prompt}

\begin{reply}[Reply 2]
\begin{lstlisting}
evaluation_plan = [
    # After action 0 (approach half-eaten apple)
    (0, {'can_grasp': ('half-eaten apple', 'side'), 
         'collision_free': (), 
         'timeout': (), 
         'check_motion_health': (),
         'can_reach': ('half-eaten apple', 'side')}, 
        (True, '', True, True, True)),
    
    # After action 1 (pick a half-eaten apple)
    (1, {'holding': (), 
         'collision_free': (), 
         'timeout': (), 
         'check_motion_health': ()}, 
        (True, '', True, True)),
    
    # After action 2 (place a half-eaten apple in the trash can)
    (2, {'at_location': ('half-eaten apple', 'large red trash can'), 
         'collision_free': (), 
         'timeout': (), 
         'check_motion_health': ()}, 
        (True, '', True, True)),
    
    # After action 3 (approach glass with yellowish liquid)
    (3, {'can_grasp': ('glass with yellowish liquid', 'top'), 
         'collision_free': (), 
         'timeout': (), 
         'check_motion_health': (), 
         'can_reach': ('glass with yellowish liquid', 'top')}, 
        (True, '', True, True, True)),
    
    # After action 4 (pick glass with yellowish liquid)
    (4, {'holding': (), 
         'collision_free': (), 
         'timeout': (), 
         'check_motion_health': ()}, 
        (True, '', True, True)),
    
    # After action 5 (drop glass in the trash can)
    (5, {'at_location': ('glass with yellowish liquid', 'large red trash can'), 
         'collision_free': (), 
         'timeout': (), 
         'check_motion_health': ()}, 
        (True, '', True, True))
]
\end{lstlisting}
\end{reply}

\begin{prompt}[Prompt 3: retuning in simulation]
The robot failed during the task\_plan's execution. The details are:\\

failure index: 2\\
failed action: place\\
current\_arguments: (`large red trash can', 0.2, 0.5, 0.03)\\
motion score: 0.015802350054063646\\
failure reason: Action failed due to: collision\_free() Collision encountered with glass with yellowish liquid |.\\
Past failures of this action:\\
FAILURE: place: large red trash can (0.2, 0.5, 0.03) | score = 0.015802350054063646\\

The score indicates the suitability of a combination. A higher score is better. First, explain how the changes you are making will improve the chances of success of the task.\\

Then alter the arguments of the failed action at index 2 in `task\_plan', to overcome the failure.\\
Do not use other actions. Make in-place change in `task\_plan'.
\end{prompt}

\begin{reply}[Reply 3]
The failure occurred because of a collision with the glass containing yellowish liquid during the placement of the half-eaten apple in the large red trash can. According to the motion score, the chosen parameters for the placement action were not suitable and led to a collision.\\

To improve the chances of success:
\begin{itemize}
    \item We can adjust the `obstacle\_clearance' argument to be larger, ensuring that the robot maintains a safer distance from surrounding objects and avoids collisions.
    \item Additionally, the robot's speed during placement can be reduced to increase precision and control, which is particularly important in a cluttered environment.
\end{itemize}
By making these adjustments, we can better ensure a collision-free motion and enhance the probability of a successful outcome. Now, let's make the necessary changes to the arguments of action 2 in the `task\_plan':
\begin{itemize}
    \item Increase `obstacle\_clearance' from 0.03 to a larger value, let's say 0.05, to provide more clearance from obstacles.
    \item Decrease the `speed' value to make the motion more controlled. Changing from 0.5 to a lower value such as 0.3, can help improve precision.
\end{itemize}

\begin{lstlisting}
# Update the task_plan with new arguments for action 2
task_plan[2] = (2, 'place', ('large red trash can', 0.2, 0.3, 0.05))
\end{lstlisting}
\end{reply}

\begin{prompt}[Prompt 4: retuning in simulation]
The robot failed during the task\_plan's execution. The details are:\\

failure index: 2\\
failed action: place\\
current\_arguments: (`large red trash can', 0.2, 0.3, 0.05)\\
motion score: 0.0308876651339363\\
failure reason: Action failed due to: at\_location((`half-eaten apple', `large red trash can')) Observed = False Expected = True | collision\_free() Collision encountered with glass with yellowish liquid |.\\
Past failures of this action:\\
FAILURE: place: large red trash can (0.2, 0.5, 0.03) | score = 0.015802350054063646\\
FAILURE: place: large red trash can (0.2, 0.3, 0.05) | score = 0.0308876651339363\\

The score indicates the suitability of a combination. A higher score is better. First, explain how the changes you are making will improve the chances of success of the task. \\

Then alter the arguments of the failed action at index 2 in `task\_plan', to overcome the failure.\\

Do not use other actions. Make in-place change in `task\_plan'.
\end{prompt}

\begin{reply}[Reply 4]
The second failure occurred again due to a collision, but also because the half-eaten apple was not placed successfully in the large red trash can. Despite the adjustments made, the placement still resulted in a collision, and the objective wasn't achieved. The increased motion score reflects the improvements, but it still wasn't enough to avoid the collision and ensure the successful placement.

To further improve the chances of success, we should consider the following changes:
\begin{itemize}
    \item Further increasing the `obstacle\_clearance' might help to avoid collisions by ensuring the robot stays further away from nearby objects. Since the trash can is likely large enough to accommodate this change, we can use this strategy without affecting the ability to place the object inside it.
    \item We could also consider modifying the `orientation' parameter to allow more flexibility in the placement. This might help to avoid collisions with the glass with yellowish liquid.
    \item Maintaining a slower speed for the placement action is a good approach as it helps in precise and controlled movements, so we will keep the speed parameter as adjusted previously (0.3).
\end{itemize}
We will increase the `obstacle\_clearance' further and adjust the `orientation' to a lower value. These combined should increase the placement accuracy and reduce the probability of collision. Let's change the \texttt{obstacle\_clearance} to 0.07 and \texttt{orientation} to 0.1.

\begin{lstlisting}
# Update the task_plan with new arguments for the failed action 2
task_plan[2] = (2, 'place', ('large red trash can', 0.1, 0.3, 0.07))
\end{lstlisting}
\end{reply}

\begin{prompt}[Prompt 5: retuning in simulation]
The robot failed during the task\_plan's execution. The details are:\\

failure index: 2\\
failed action: place\\
current\_arguments: (`large red trash can', 0.1, 0.3, 0.07)\\
motion score: 0.04319835434592172\\
failure reason: Action failed due to: at\_location((`half-eaten apple', `large red trash can')) Observed = False Expected = True |collision\_free() Collision encountered with glass with yellowish liquid |.\\
Past failures of this action:\\
FAILURE: place: large red trash can (0.2, 0.5, 0.03) | score = 0.015802350054063646\\
FAILURE: place: large red trash can (0.2, 0.3, 0.05) | score = 0.0308876651339363\\
FAILURE: place: large red trash can (0.1, 0.3, 0.07) | score = 0.04319835434592172\\

The score indicates the suitability of a combination. A higher score is better. First, explain how the changes you are making will improve the chances of success of the task.\\

Then alter the arguments of the failed action at index 2 in `task\_plan', to overcome the failure.\\

Do not use other actions. Make in-place change in `task\_plan'.
\end{prompt}

\begin{reply}[Reply 5]
The continuous failures due to collisions and unsuccessful placement suggest that the adjustments made so far are insufficient in preventing collisions with the glass containing yellowish liquid. Since the motion score improved with each change but still resulted in a collision, the modifications to orientation and obstacle clearance did not fully resolve the issue.\\

Taking into account the nature of the failures and also the fact that the collision happens with the same object (glass with yellowish liquid), it's possible that the spatial arrangement of the objects on the table is causing the robot to hit the glass while attempting to place the half-eaten apple in the trash can, which could mean that the robot's arm trajectory intersects with the glass's position. To improve the chances of success, we now have to find a different strategy within the constraints of the available action arguments that might involve rethinking the approach towards the trash can.\\

To overcome this issue, one strategy could be to:\\
\begin{itemize}
\item Further decrease the `orientation' value, allowing even more flexibility in how the apple is placed inside the trash can in hopes of avoiding a collision.
\item Further increase the `obstacle\_clearance' to maximize the distance from the glass with yellowish liquid. However, we must be cautious not to set the clearance too high, as it might make it impossible for the robot to reach into the trash can.
\item If the previous adjustments to speed and clearance did not affect the result, we may consider slightly increasing the speed again to explore if a different arm trajectory is selected by the robot's planning system which avoids the glass.
\end{itemize}

Let's try:
\begin{itemize}
\item Maintaining the slower speed to assist with careful placement: 0.3 (As lower speed didn't seem to be the issue.)
\item Increasing the `obstacle\_clearance' to 0.08 to see if a larger clearance can help avoid the glass.
\item Setting the `orientation' to 0, giving the robot maximum flexibility in how the apple is placed inside the trash can.
\end{itemize}

Since the motion scores do seem to be sensitive to the changes in the `obstacle\_clearance' and `orientation', these changes could potentially avoid the collision and allow successful placement. Let's implement these in the `task\_plan':

\begin{lstlisting}
# Update the task_plan with new arguments for the failed action 2
task_plan[2] = (2, 'place', ('large red trash can', 0, 0.3, 0.08))
\end{lstlisting}
\end{reply}

\begin{prompt}[Prompt 6: replanning in simulation]
The robot failed during the task\_plan's execution. The details are:\\
failure index: 2\\
failed action: place\\
current\_arguments: (`large red trash can', 0, 0.3, 0.08)\\
motion score: 0.01331668352320572\\
failure reason: Action failed due to: collision\_free() Collision encountered with glass with yellowish liquid |.\\

Past failures:\\
FAILURE: place: large red trash can(0.2, 0.5, 0.03) | score = 0.015802350054063646\\
FAILURE: place: large red trash can(0.2, 0.3, 0.05) | score = 0.0308876651339363\\
FAILURE: place: large red trash can(0.1, 0.3, 0.07) | score = 0.04319835434592172\\
FAILURE: place: large red trash can(0, 0.3, 0.08) | score = 0.01331668352320572\\

Past successes:\\
SUCCESS: approach: half-eaten apple(0.5, 0.03, `side') | score = 0.002667235367929972\\
SUCCESS: approach: half-eaten apple(0.5, 0.03, `side') | score = 0.002641962213308816\\
SUCCESS: approach: half-eaten apple(0.5, 0.03, `side') | score = 0.00390897408684617\\
SUCCESS: approach: half-eaten apple(0.5, 0.03, `side') | score = 0.002450670636082305\\
SUCCESS: pick: half-eaten apple(0.5, 0.03, `side') | score = 0.03308374125099294\\
SUCCESS: pick: half-eaten apple(0.5, 0.03, `side') | score = 0.031066406658904498\\
SUCCESS: pick: half-eaten apple(0.5, 0.03, `side') | score = 0.04395666157923764\\
SUCCESS: pick: half-eaten apple(0.5, 0.03, `side') | score = 0.03289765198915318\\

Perform replanning by either using alternative action functions or altering the object interaction sequence.\\
Output the code as you did at first by assigning the full task plan to the variable `task\_plan' in a single statement.\\
Take into account the parameters I have asked you to change during our conversation if any.\\
In the new plan, make sure it is within plus or minus 5 actions of the original task plan.
\end{prompt}

\begin{reply}[Reply 6]
Given the repeated failures at placing the half-eaten apple into the large red trash can due to collisions with the glass containing yellowish liquid, an alternative approach should be considered. Instead of trying to place the half-eaten apple directly into the trash can and risking another collision, the robot should drop the apple into the trash can. The drop function does not require the robot to be as close to the container as the place function, possibly allowing the robot to avoid the glass.\\

Additionally, the sequence of interacting with the objects should be changed to handle the glass with yellowish liquid first, moving it out of the way to prevent further collisions when the robot attempts to place or drop the half-eaten apple into the trash can.\\

Let's make the following changes to the task plan:
\begin{enumerate}
\item The robot will approach the glass with yellowish liquid first.
\item Then, the robot will pick up the glass with yellowish liquid.
\item The robot will drop the glass into the large red trash can.
\item The robot will then approach the half-eaten apple.
\item Next, the robot will pick up the half-eaten apple.
\item Finally, the robot will drop the half-eaten apple into the large red trash can.
\end{enumerate}

This adjusted plan should avoid the immediate collision issue and will comply with the requirement to stay within plus or minus 5 actions of the original task plan. Here's the new plan:

\begin{lstlisting}
# Define the new task plan
task_plan = [
# Approach the glass with yellowish liquid first with a moderate speed
# and obstacle clearance
(0, 'approach', ('glass with yellowish liquid', 0.5, 0.03, 'top')),

# Pick up the glass with yellowish liquid at a moderate speed ensuring a top grasp
(1, 'pick', ('glass with yellowish liquid', 0.5, 0.03, 'top')),

# Drop the glass with yellowish liquid in the large red trash can
(2, 'drop', ('large red trash can', 0.5, 0.03)),

# Now approach the half-eaten apple with moderate speed and obstacle clearance
(3, 'approach', ('half-eaten apple', 0.5, 0.03, 'side')),

# Pick up the half-eaten apple at a moderate speed ensuring a side grasp
(4, 'pick', ('half-eaten apple', 0.5, 0.03, 'side')),

# Drop the half-eaten apple in the large red trash can
(5, 'drop', ('large red trash can', 0.5, 0.03))
]
\end{lstlisting}
\end{reply}

\begin{prompt}[Prompt 7: generating new evaluation plan]
The robot may not be able to execute an action function or encounter object collision during execution. Thus, it is required to check for completion of each action function after they have been performed.\\

For this, we define some `checking functions' written in Python. These checking functions, which can be imported from the `checking\_functions' library are:\\

can\_grasp(object\_to\_grasp: str, grasp: str) $->$ bool: \# Returns True if the robot is close enough to the `object\_to\_grasp' to securely grasp it with the determined grasp `side' or `top'\\

holding() $->$ bool \# Returns True if the robot is holding an object\\

at\_location(object: str, location: str) $->$ bool: \# Returns True if the `object' is at the `location'\\

collision\_free() $->$ str: \# If the robot encounters a collision while executing the preceding action, returns the object label string. Otherwise, `'.\\

timeout() $->$ Bool: \# Returns True if the preceding action was executed in a timely fashion\\

check\_motion\_health() $->$  bool: \# Returns True if the robot's motion during the preceding action was safe for its hardware\\

can\_reach(goal: str, grasp: str) $->$ bool: \# Returns True if it is feasible for the robot to reach the `goal' object or location from the current state from the side determined by the grasp argument `side' or `top'. Objects that are out of the workspace will always return false.\\

The grasp argument is the same as the one in the `approach' and `pick' functions. It assumes one of the two values {`top', `side'}\\

Using the `checking\_functions', `locations', and `objects', define an evaluation plan (named `evaluation\_plan') to verify the successful execution of each action. Additionally, for each action verify without fail:
\begin{itemize}
\item collision-free
\item timely motion
\item motion health
\end{itemize}

Output this plan as a Python list of tuples, where each tuple is of the form (action number int, dictionary with `check\_function' names as keys and a tuple of arguments as value, tuple of expected outputs). Do not assume any other object or location, beyond those in `object\_labels'.\\

Each tuple is meant to be checked after the action with the corresponding number.\\

Generate the entire plan. No reasoning, direct output.
\end{prompt}

\begin{reply}[Reply 7]
\begin{lstlisting}
evaluation_plan = [
# After action 0 (approach glass with yellowish liquid)
(0, {'can_grasp': ('glass with yellowish liquid', 'top'),
'collision_free': (),
'timeout': (),
'check_motion_health': (),
'can_reach': ('glass with yellowish liquid', 'top')},
(True, '', True, True, True)),

# After action 1 (pick glass with yellowish liquid)
(1, {'holding': (),
'collision_free': (),
'timeout': (),
'check_motion_health': ()},
(True, '', True, True)),

# After action 2 (drop glass with yellowish liquid into trash can)
(2, {'at_location': ('glass with yellowish liquid', 'large red trash can'),
'collision_free': (),
'timeout': (),
'check_motion_health': ()},
(True, '', True, True)),

# After action 3 (approach half-eaten apple)
(3, {'can_grasp': ('half-eaten apple', 'side'),
'collision_free': (),
'timeout': (),
'check_motion_health': (),
'can_reach': ('half-eaten apple', 'side')},
(True, '', True, True, True)),

# After action 4 (pick a half-eaten apple)
(4, {'holding': (),
'collision_free': (),
'timeout': (),
'check_motion_health': ()},
(True, '', True, True)),

# After action 5 (drop half-eaten apple into the trash can)
(5, {'at_location': ('half-eaten apple', 'large red trash can'),
'collision_free': (),
'timeout': (),
'check_motion_health': ()},
(True, '', True, True))
]
\end{lstlisting}
\end{reply}

\begin{prompt}[Prompt 8: retuning in simulation]
The robot failed during the task\_plan's execution. The details are:\\

failure index: 3\\
failed action: approach\\
current\_arguments: (`half-eaten apple', 0.5, 0.03, `side')\\
motion score: 0.027770374130814297\\
failure reason: Action failed due to: collision\_free() Collision encountered with half-eaten apple |.\\
Past successes of this action:\\
SUCCESS: approach: half-eaten apple (0.5, 0.03, `side') | score = 0.002667235367929972\\
SUCCESS: approach: half-eaten apple (0.5, 0.03, `side') | score = 0.002641962213308816\\
SUCCESS: approach: half-eaten apple (0.5, 0.03, `side') | score = 0.00390897408684617\\
SUCCESS: approach: half-eaten apple (0.5, 0.03, `side') | score = 0.002450670636082305\\

The score indicates the suitability of a combination. A higher score is better. First, explain how the changes you are making will improve the chances of success of the task.\\

Then alter the arguments of the failed action at index 3 in `task\_plan', to overcome the failure.\\

Do not use other actions. Make in-place change in `task\_plan'.
\end{prompt}

\begin{reply}[Reply 8]
The failure during the approach action towards the half-eaten apple suggests that despite previous successful attempts with the same parameters, a collision with the half-eaten apple occurred, possibly due to a slight change in the apple's position or the robot's starting position.\\

To improve the chances of successfully approaching the half-eaten apple without collision, we can:
\begin{enumerate}
\item Increase the `obstacle\_clearance' parameter to give the robot a wider berth around the apple, reducing the risk of bumping into it.
\item Adjust the speed slightly lower to provide the robot with more control during its approach, which could allow for finer adjustments to be made if the robot detects a potential collision.
\end{enumerate}

These adjustments are practical and should give the robot a better chance of avoiding collision with the half-eaten apple or any other nearby objects. Let's increase the `obstacle\_clearance' to 0.05 meters and decrease the speed to 0.4.\\

Below is the updated action 3 in the `task\_plan':

\begin{lstlisting}
# Update the task_plan with new arguments for the failed action 3
task_plan[3] = (3, 'approach', ('half-eaten apple', 0.4, 0.05, 'side'))
\end{lstlisting}
\end{reply}

\begin{prompt}[Prompt 9: retuning in simulation]
The robot failed during the task\_plan's execution. The details are:\\

failure index: 3\\
failed action: approach\\
current\_arguments: (`half-eaten apple', 0.4, 0.05, `side')\\
motion score: 0.028076367878740724\\
failure reason: Action failed due to: collision\_free() Collision encountered with half-eaten apple |.\\
Past failures of this action:\\
FAILURE: approach: half-eaten apple (0.4, 0.05, `side') | score = 0.028076367878740724\\
Past successes of this action:\\
SUCCESS: approach: half-eaten apple (0.5, 0.03, `side') | score = 0.002667235367929972\\
SUCCESS: approach: half-eaten apple (0.5, 0.03, `side') | score = 0.002641962213308816\\
SUCCESS: approach: half-eaten apple (0.5, 0.03, `side') | score = 0.00390897408684617\\
SUCCESS: approach: half-eaten apple (0.5, 0.03, `side') | score = 0.002450670636082305\\
SUCCESS: approach: glass with yellowish liquid (0.5, 0.03, `top') | score = 0.030006371189952436\\

The score indicates the suitability of a combination. A higher score is better. First, explain how the changes you are making will improve the chances of success of the task.\\

Then alter the arguments of the failed action at index 3 in `task\_plan', to overcome the failure.\\

Do not use other actions. Make in-place change in `task\_plan'.
\end{prompt}

\begin{reply}[Reply 9]
The motion score indicates that the suitability of the action has improved slightly by increasing the `obstacle\_clearance' and reducing the speed, compared to the first attempt, but the robot still encountered a collision with the half-eaten apple. Given the failure despite these adjustments, a better approach may be to change the `grasp' parameter from `side' to `top', which could allow the robot to approach from a different angle, potentially avoiding the collision.\\

By changing the grasp type to `top', it's possible that the robot changes its approach trajectory to be more vertical, which can be helpful particularly if the `side' approach was causing the collision. This top-down approach might help the robot avoid colliding with the half-eaten apple or any other surrounding objects on the table.\\

Let's implement these changes in the `task\_plan'. The `speed' will remain at 0.4, which is slower than the initial value and will allow the robot to maintain better control over its movement. The `obstacle\_clearance' will remain at 0.05, as this value offered an improvement. The `grasp' parameter will change to `top'.\\

Here is the updated action 3 in the `task\_plan':

\begin{lstlisting}
# Update the task_plan with new arguments for the failed action 3
task_plan[3] = (3, 'approach', ('half-eaten apple', 0.4, 0.05, 'top'))
\end{lstlisting}
\end{reply}

\end{document}